\documentclass[10pt,twocolumn,letterpaper]{article}

\usepackage{cvpr}              % To produce the CAMERA-READY version

\definecolor{cvprblue}{rgb}{0.21,0.49,0.74}
\usepackage[pagebackref,breaklinks,colorlinks,allcolors=cvprblue]{hyperref}

\newif\ifmodify 
\modifytrue % CARE: comment this line to compile the final version
\ifmodify

\else

\fi
\usepackage{comment}
\usepackage{multirow}
\usepackage{colortbl}
\usepackage{graphicx}
\usepackage{amsmath} 
\usepackage{booktabs}       % professional-quality tables
\usepackage{amsfonts}       % blackboard math symbols
\usepackage{nicefrac}       % compact symbols for 1/2, etc.
\usepackage{microtype}      % microtypography
\usepackage{xcolor}         % colors
\usepackage{graphicx}
\definecolor{light-gray}{gray}{0.92}

\def\paperID{8525} % *** Enter the Paper ID here
\def\confName{CVPR}
\def\confYear{2025}

\title{
QuartDepth: Post-Training \underline{Qua}ntization for \\ \underline{R}eal-\underline{T}ime Depth Estimation on the Edge
}

\author{
Xuan Shen$^1$, Weize Ma$^2$, Jing Liu$^3$, Changdi Yang$^1$, Rui Ding$^2$, Quanyi Wang$^4$, \\
Henghui Ding$^5$, Wei Niu$^6$, Yanzhi Wang$^1$, Pu Zhao$^{1}$\thanks{Corresponding Author}, Jun Lin$^{2}$\footnotemark[1], Jiuxiang Gu$^{7}$\footnotemark[1] \\
$^1$Northeastern University,
$^2$Nanjing University,
$^3$Monash University, \\
$^4$Nanjing University of Information Science and Technology, 
$^5$Fudan University, \\
$^6$University of Georgia,
$^7$Adobe Research \\
{
\tt\small \{shen.xu,yanz.wang,p.zhao\}@northeastern.edu, 
weizema@smail.nju.edu.cn,
}
\\
{
\tt\small 
jlin@nju.edu.cn, 
jigu@adobe.com
}
}

\begin{document}

\maketitle

\begin{abstract}

Monocular Depth Estimation (MDE) has emerged as a pivotal task in computer vision, supporting numerous real-world applications.
% including autonomous driving, robotics, augmented reality, and scene reconstruction.
However, deploying accurate depth estimation models on resource-limited edge devices, especially Application-Specific Integrated Circuits (ASICs), is challenging due to the high computational and memory demands.
Recent advancements in foundational depth estimation deliver impressive results but further amplify the difficulty of deployment on ASICs. 
To address this, we propose \textbf{QuartDepth} which adopts post-training quantization to quantize MDE models with hardware accelerations for ASICs. 
%optimize and accelerate MDE models specifically for ASICs. 
Our approach involves quantizing both weights and activations to 4-bit precision, reducing the model size and computation cost. 
To mitigate the performance degradation,
% typically associated with aggressive quantization, 
we introduce activation polishing and compensation algorithm applied before and after activation quantization, as well as a weight reconstruction method for minimizing errors in weight quantization.
Furthermore, we design a flexible and programmable hardware accelerator by supporting kernel fusion and customized instruction programmability, enhancing throughput and efficiency.
Experimental results demonstrate that our framework achieves competitive accuracy while enabling fast inference and higher energy efficiency on ASICs, bridging the gap between high-performance depth estimation and practical edge-device applicability.
% This research contributes a scalable solution for deploying depth estimation models in latency-sensitive applications across various industries.
Code: \url{https://github.com/shawnricecake/quart-depth}
\end{abstract}

\section{Introduction}

% Background
Monocular Depth Estimation (MDE) is a critical task in computer vision, essential for a wide range of applications including robotics~\cite{icra_2019_fastdepth}, autonomous driving~\cite{wang2019pseudo, you2020pseudo}, virtual reality~\cite{rasla2022relative}, and 3D reconstruction~\cite{ranftl2020towards, Wei2021CVPR, zhang2022hierarchical}. The objective of MDE is to estimate depth information from a single image, making it particularly valuable for scenarios where stereo or multi-view depth sensors are impractical. 
Previously, depth estimation methods have been classified into three main approaches: learning metric depth~\cite{yuan2022neural, yin2021virtual, yang2021transformer, bhat2021adabins}, learning relative depth~\cite{chen2016single, chen2020oasis, 8578138, Xian_2020_CVPR}, and learning affine-invariant depth~\cite{yin2022towards, yin2021learning, ranftl2021vision, ranftl2020towards, zhang2022hierarchical}. 
Among these, metric depth methods have demonstrated remarkable performance across benchmarks but often show limited generalizability to diverse, real-world images, which is compounded by variations in camera parameters between training and test sets.

\begin{figure}[t]
  \centering
  \includegraphics[width=1.0\linewidth]{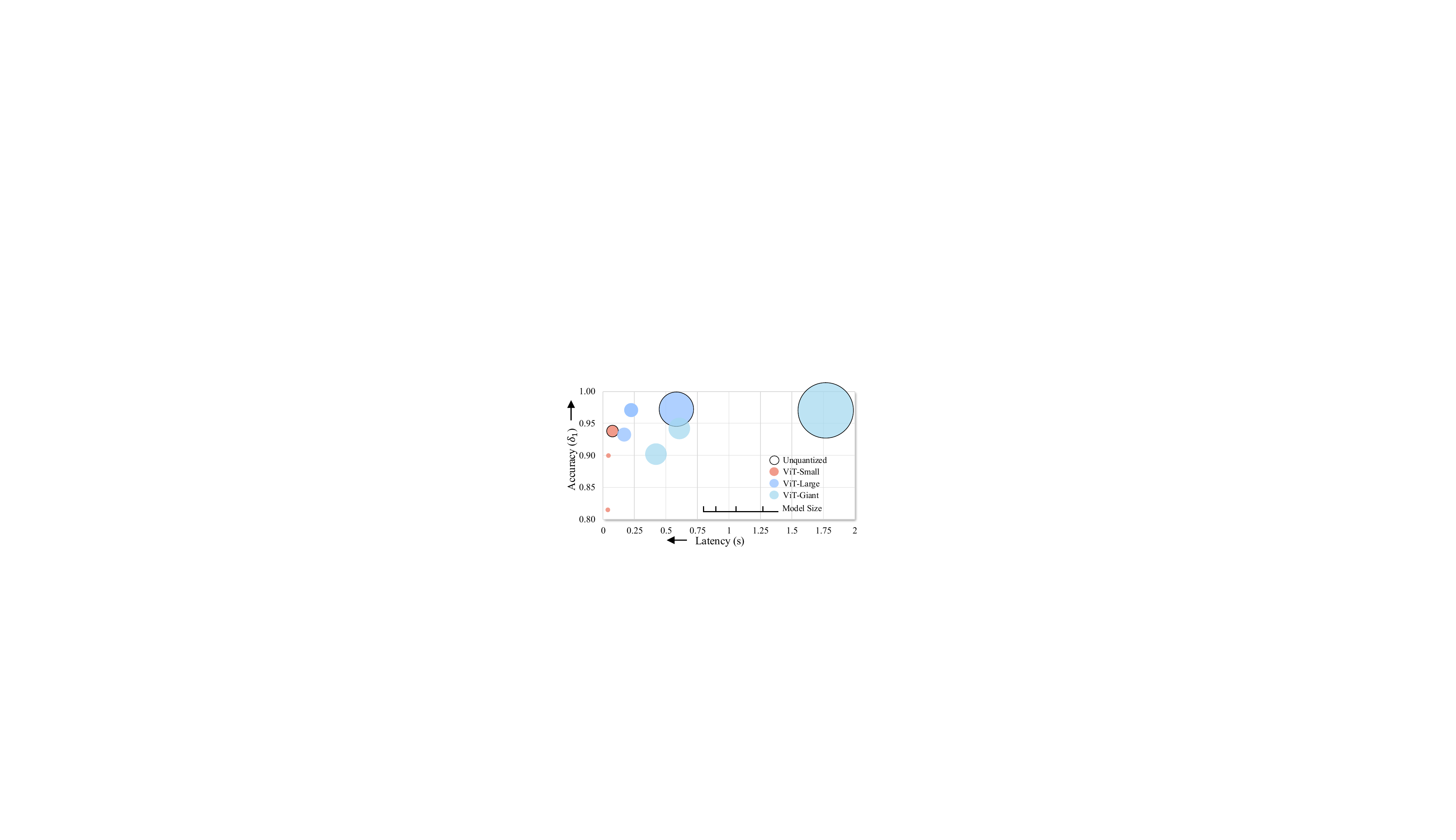}
  \caption{
  Visualization of accuracy vs. latency on ASIC for monocular depth estimation models with different backbones.
  % Visualization of latency versus accuracy, with bubble size representing the model size.
  % Accuracy is evaluated with Metric3D~\cite{yin2023metric3d} on NYUv2~\cite{nyuv2} dataset. 
  % Latency is tested in 256x256 resolution on 
  % emulator of dedicated hardware implementation.
  }
  \label{fig:first_page_figure}
\end{figure}

% Motivation
Recently, inspired by the rise of “foundation models”~\cite{foundation_model_trendency,zhao2024fully}, a new wave of MDE models~\cite{yin2023metric3d, depth_anything_v1, zoedpeth, 9316778} has been developed. 
These foundational MDE models exhibit superior generalization and robustness, effectively extending their applicability to in-the-wild images and accommodating a wide array of camera parameters. 
% This capability marks a significant step forward, enabling the production of high-quality depth maps for images captured under varying real-world conditions.
However, while these foundational MDE models achieve impressive performance, they rely on sophisticated and computationally intensive algorithms. 
This reliance on complex architectures and operations significantly increases computational costs, making such models resource-heavy and challenging to deploy on edge devices or specialized hardware such as Application-Specific Integrated Circuits (ASICs)~\cite{7460664, icra_2019_fastdepth, wofk2019fastdepth, poudel2019fast}. 
The high computational demands also hinder real-time performance and scalability, which are critical for practical applications such as autonomous navigation~\cite{badue2021self}, robotic perception~\cite{WEN2022255}, and augmented reality~\cite{Suri2023}. 
This constraint underscores the need for more efficient approaches that can maintain high accuracy while reducing computational overhead. 
Among common techniques like pruning~\cite{icra_2019_fastdepth, elkerdawy2019lightweight,zhang2022advancing, shen2024numerical}, knowledge distillation~\cite{garg2016unsupervised, pilzer2019refine}, and architecture optimization~\cite{Alhashim2018,zhan2021achieving,wu2022compiler,rtseg,li2022pruning, shen2024lazydit}, which mainly reduce model complexity, quantization ~\cite{percentile, li2021brecq, adaround, wei2022qdrop} stands out for achieving both compactness and speed. 
Quantization~\cite{wu2020integer, ding2022towards, li2023repq, yuan2021ptq4vit, lin2022fqvit} reduces weights and activations to lower bit-widths (e.g., 4-bit or 8-bit integers), minimizing model size while accelerating inference through optimized integer kernels specifically designed for edge devices. 
This dual benefit makes it ideal for deploying high-performing models on resource-limited edge hardware. 
Given the complexity of large foundational MDE models and their extensive datasets~\cite{segment_anything, bdd100k, open_images_dataset_v4}, we employ Post-Training Quantization (PTQ)~\cite{yuan2021ptq4vit,li2021brecq,lin2022fqvit,zhao-etal-2024-pruning,shen2024search}  to facilitate efficient deployment without retraining.

% Our method
In this paper, we propose \textbf{QuartDepth}, a post-training quantization framework designed for real-time depth estimation on edge devices. 
We begin by analyzing the outlier deviant distribution in MDE foundation models~\cite{yin2023metric3d, depth_anything_v1}. 
Through an in-depth per-channel analysis of these abnormal distributions, we identify persistent extreme outliers in the depth decoders. 
To address this, we propose the LogNP polishing optimization method to smooth the outliers and transform the abnormal distribution into a more quantization-friendly, normalized distribution.
% Additionally, we find that the attention mechanism causes significant quantization errors due to the difficulty in quantizing post-softmax distributions, which we address by introducing the adaptive \textit{Log2} strategy.
Furthermore, to mitigate the loss from activation quantization, we employ a compensation algorithm that updates the weights to minimize the activation quantization error.
Next we perform weight quantization with a weight reconstruction method that leverages gradients to minimize second-order weight quantization errors.
% Meanwhile, we develop specialized computational kernels on ASIC to facilitate the inference acceleration of quantized MDE models. We propose a flexible and programmable hardware accelerator architecture under which the matrix multiplication and nonlinear computations of the MDE model can be significantly accelerated. We have designed corresponding hardware implementations for matrix multiplication calculations targeting floating-point numbers, W8A8, W4A4, and W4A8, and nonlinear activation function computations are also rapidly computed on hardware while maintaining a very high level of precision. 
% \todo{todo: @weize, please continue to write about the hardware contribution from here.}
% Meanwhile, we propose a flexible and programmable hardware accelerator architecture significantly speeds up our proposed quantization framework for MDE model. 
% Our hardware design supports efficient matrix multiplication for floating-point precision, W8A8, W4A4, and W4A8 configurations. 
% The proposed hardware architecture also supports the \textbf{Quart-Depth} algorithm with almost no latency overhead.
% \todo{todo: @weize, do we need to include the nonlinear activation accelerations in this paper? need to refine this part, we need to focus on ASIC noval design}
Meanwhile, we design a novel flexible and programmable hardware accelerator by supporting kernel fusion and customized instruction programmability, which enables direct processing of intermediate results and concurrent execution of computational tasks.
In detail, we design specialized computation kernels for W4A4 and W4A8 configurations, which enables full utilization of external memory bandwidth.
Also, we design a novel programmable vector computation array to support the proposed LogNP polishing optimization, effectively hiding its additional overhead.
% \todo{@weize, please check if above is right:}

In summary, our contributions are outlined as follows,

\begin{itemize}[label={}, leftmargin=*]

\item \textbf{1.} We observe a challenging outlier deviant distribution for activations  and propose the LogNP polishing for activation quantization, which transforms these outliers into a quantization-friendly, normalized distribution.

\item \textbf{2.} For weight quantization, we first update the weights to compensate the error of activation quantization, and then quantize the updated weights with weight reconstruction to minimize the second-order weight quantization error.

% \item \textbf{3.} We develop a hardware architecture for quantized MDE models that can efficiently perform matrix multiplications with various numerical formats with the support of \textbf{Quart-Depth} quantization algorithm.
% \todo{todo: @weize, please focus on the hardware design}

\item \textbf{3.} We develop a novel flexible and programmable hardware accelerator on ASICs corresponding to our proposed quantization methods.

\item \textbf{4.} Comprehensive experiments confirm the effectiveness of our QuartDepth framework with superior accuracy, real-time inference and higher power efficiency on ASICs.

\end{itemize}

\section{Related Work}

\subsection{Efficient Depth Estimation}

Depth estimation from a single color image has been a prominent research focus in the field of computer vision for over a decade. 
This task plays a critical role in various applications, including robotic perception~\cite{icra_2019_fastdepth,MOSE,MeViS}, autonomous driving~\cite{wang2019pseudo, you2020pseudo}, virtual reality~\cite{rasla2022relative}, and 3D reconstruction~\cite{ranftl2020towards, Wei2021CVPR, zhang2022hierarchical}, all of which require rapid and accurate depth perception. 
The demand for real-time processing in these applications highlights the necessity for efficient model deployment. 
To meet these requirements, advanced optimization techniques such as pruning~\cite{icra_2019_fastdepth, elkerdawy2019lightweight,zhan-etal-2024-rethinking-token,shen2025sparse,liu2025toward}, knowledge distillation~\cite{garg2016unsupervised, pilzer2019refine}, and architecture optimization~\cite{Alhashim2018,yang2023pruning,zhao-etal-2024-pruning} have been adopted to compress and accelerate the model.
However, previous research has primarily focused on relatively smaller models, which, while effective, do not match the scale and capability of foundational MDE models~\cite{yin2023metric3d, depth_anything_v1} . 
These modern foundational MDE models~\cite{zoedpeth, 9316778} have gained significant attention for their ability to achieve state-of-the-art performance and generalize effectively across diverse real-world scenarios. 
By leveraging massive datasets and advanced architectures, these models provide robust depth estimation  at the cost of increased computational complexity and larger model sizes.
Thus, there is an urgent need for new model compression and acceleration techniques specifically designed to meet real-time requirements on resource-limited edge devices.

\subsection{Hardware Design}

Deploying large foundational models on edge devices presents significant challenges due to stringent constraints on computation and memory access~\cite{Surianarayanan2023ASO}.
To tackle these issues, various optimization techniques~\cite{shen2024agile,shen2024search,shen2024edgeqat,zhao-etal-2024-pruning,hotaq,zhan2024fast,zhan2024exploring,wu2024aye} have been explored to reduce model size and computational overhead. 
However, models subjected to such aggressive optimizations often fail to achieve peak performance on general-purpose CPUs and GPUs~\cite{vitcod}. 
In contrast, ASICs, optimized for specific models and algorithms, offer significant performance and energy efficiency gains, making them ideal for deploying large foundational models on edge devices~\cite{aaa, ELSA, DOTA, Approximate, Sanger, vitcod}.
% Recent works~\todo{@weize, please add some citations for works} have consistently highlighted the advantages of ASICs in this context. 
% Early efforts, such as Google's TPU~\cite{tpu}, pioneered the use of int8 low-precision computation as a mainstream optimization in deep learning, demonstrating significant improvements in inference speed and energy efficiency. 
% Similarly, $A^3$~\cite{aaa} accelerates the attention mechanism through hardware specialization and algorithmic approximation, significantly enhancing both performance and energy efficiency. 
% Early work, such as Google's TPU~\cite{tpu}, introduced int8 low-precision computation, significantly improving inference speed and energy efficiency.
Early work, such as TPU~\cite{tpu}, DNA~\cite{DNA}, and Eyeriss~\cite{eyeriss}, significantly improved inference speed and energy efficiency by focusing on specialized hardware designs tailored to the needs of deep learning models. 
% Likewise, $A^3$~\cite{aaa} enhances attention mechanisms through hardware specialization and algorithmic approximation, boosting performance and efficiency.
% More recent advancements include AccelTran~\cite{acceltran}, which introduces an ASIC architecture optimized for its dynamic inference scheme; DynaTran~\cite{tuli2023acceltran}, efficiently pruning activations at runtime with minimal overhead; and ViTCoD~\cite{vitcod}, which designs a dedicated accelerator to handle both dense and sparse workloads simultaneously, thus improving hardware utilization. 
% Recent advancements include AccelTran~\cite{acceltran}, which features an ASIC architecture optimized for dynamic inference schemes, ViTCoD~\cite{vitcod}, which incorporates a specialized accelerator to handle both dense and sparse workloads, enhancing hardware utilization, $A^3$~\cite{aaa} enhances attention mechanisms through hardware specialization and algorithmic approximation, boosting performance and efficiency.
Recent works~\cite{acceltran, FLAT, DIANA, 8k, 5nm} have proposed dedicated hardware architectures in ASICs to accelerate neural network inference.
These ASIC architectures have significantly improved inference latency and energy efficiency through solutions such as quantization, sparsity, data flow reconstruction, and algorithm-hardware co-design~\cite{codesign}.
Thus, to maximize potential of large foundational models, the advantages offered by ASICs make them platform of choice.

% \todo{@weize, please write about related work for ASIC}

\section{Analysis and Motivation}

\subsection{Latency Profiling}
\label{sec:latency_profiling}

\begin{figure}[t]
  \centering
  \includegraphics[width=1.0\linewidth]{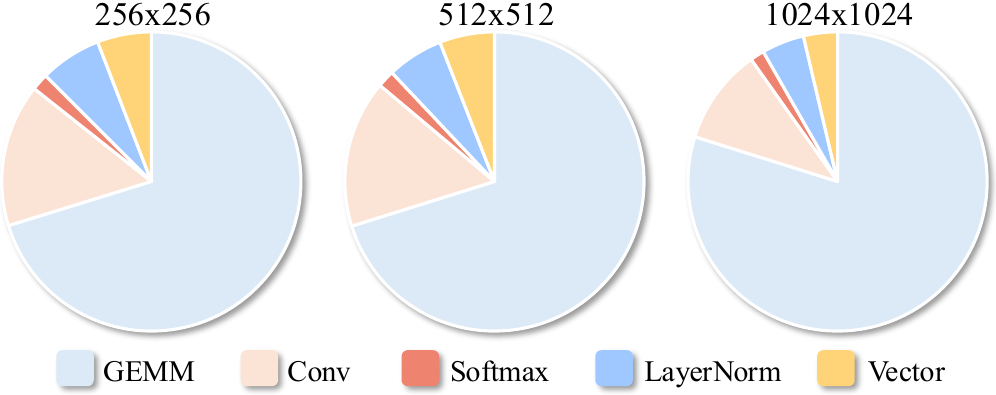}
  \caption{
  Latency profliling for 3 different resolutions using Metric3D with ViT-Large backbone. Vector includes activation functions (e.g. ReLU)
  % (e.g. SiLU, ReLU, GeLU) 
  and element-wise addition and multiplication.}
  \label{fig:analysis_profiling}
\end{figure}

To optimize edge device inference, we first profile the Float32 model to identify  the most time-consuming components for targeted optimization.
In detail, we implement the Metric3D~\cite{yin2023metric3d} model with ViT-Large~\cite{vit} backbone on a hardware performance emulator with our ASIC design. 
%~\todo{@weize, please 1. replace 'edge devices' with your testing bed and 2. write some details of profiling implementation, 3. maybe u can change the backbone size to meet our conclusion below.}
The latency profiling results in Figure~\ref{fig:analysis_profiling} show that the matrix multiplication and convolution operations dominate the inference time, while non-linear operations like softmax and layer normalization contribute minimally. 
The vector operations include activation functions such as SiLU, ReLU, and GeLU, as well as element-wise addition and multiplication, which also contribute minimally to the overall latency.
This motivates our focus on quantization and ASIC design for linear and convolution layers, keeping non-linear operations in floating point to preserve accuracy.

\subsection{Deviant Distribution}\label{sec:analysis_distribution}

\begin{figure}[t]
  \centering
  \includegraphics[width=1.0\linewidth]{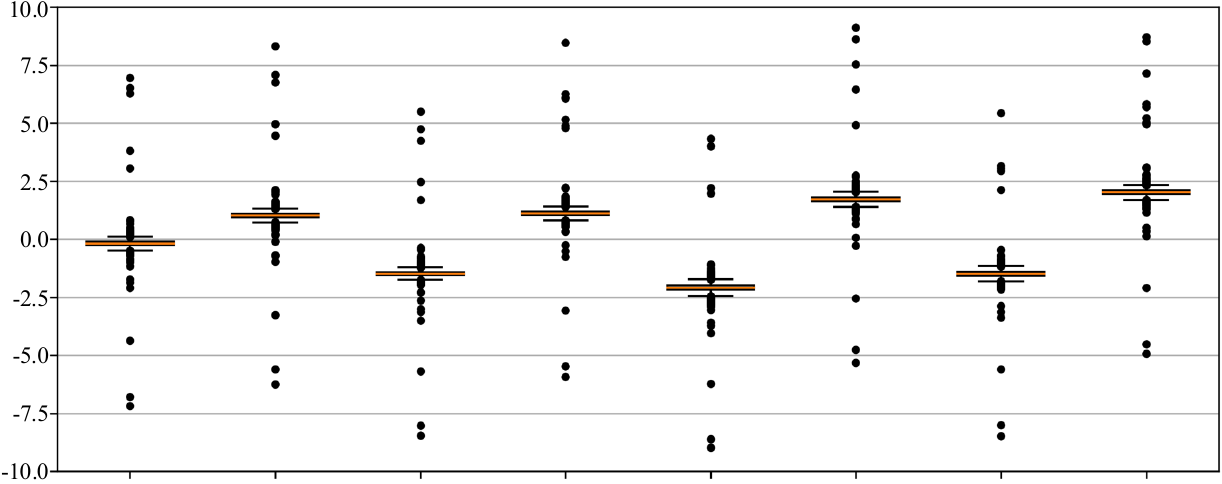}
  \caption{
  Visualization of outliers in various channels at the decoder~\cite{depth_decoder} of Metric3D model on NYUv2 dataset.
  }
  \label{fig:analysis_outlier_boxplot}
\end{figure}

\begin{figure}[h]
  \centering
  \includegraphics[width=1.0\linewidth]{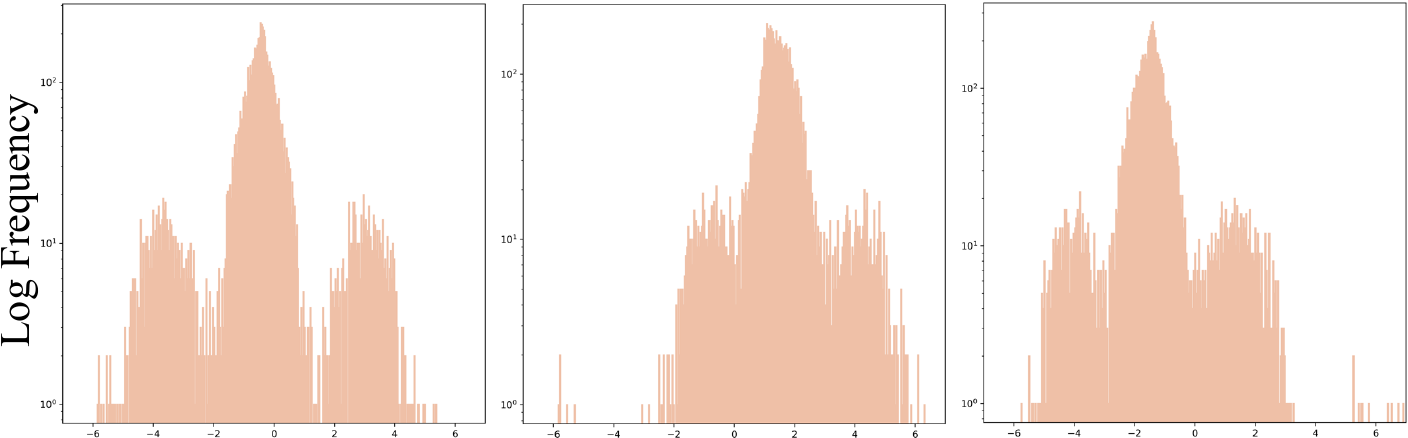}
  \caption{
  Visualization for the activation data distribution of different channels in decoder with $\log$ frequency.
  }
  \label{fig:analysis_outlier_distribution}
\end{figure}

We investigate the deviant distribution in MDE models from a per-channel perspective.
In Figure~\ref{fig:analysis_outlier_boxplot}, we visualize the outliers across various channels of an activation in the decoder~\cite{depth_decoder} of the Metric3D model on NYUv2 dataset.
We observe large amount of outliers and significant variability in outlier values across different channels, making per-tensor quantization challenging. 
This motivates the adoption of a per-channel quantization method to better handle these variations.
However, for per-channel quantization, the presence of outliers that deviate significantly from the main data distribution, as shown in Figure~\ref{fig:analysis_outlier_distribution}, still leads to substantial quantization errors.
Thus, smoothing these large outliers and integrating them into the main data distribution is crucial for maintaining model accuracy.

\section{Quantization Framework}

We first  quantize the polished activations with LogNP polishing  (Section~\ref{sec:activation_polishing}). Then we update weights to compensate the loss of activation quantization (Section~\ref{sec:compensation}). Finally, we quantize updated weights  (Section~\ref{sec:weight_quant}).

\subsection{Preliminary}
\label{sec:preliminary}

\textbf{Quantization.}
We mainly adopt the hardware-efficient PTQ methods.
For uniform quantization, the quantize and dequantize operations for $ x $ can be defined as follows,
\begin{equation}
     x_q = \text{CLIP} \left( \left\lfloor \frac{x}{s} \right\rceil + zp, 0, 2^k - 1 \right) ,
\end{equation}
\begin{equation}
\hat{x} = s \cdot ( x_q - z) ,
\end{equation}
where $s$ represents the scale and $zp$ is the zero point. The operator $\lfloor \cdot \rceil$ indicates rounding to the nearest integer, and the CLIP function constrains values that fall outside the range of a $k$-bit integer.

As for Log2 quantization, which is mainly used for post-softmax quantization, can be defined as follows,
\begin{equation}
x_q = \text{CLIP} \left( \left\lfloor -\log_2 \frac{x}{s} \right\rceil, 0, 2^k - 1 \right),
\end{equation}
\begin{equation}
\hat{x} = s \cdot 2^{-x_q}.
\end{equation}

\noindent
\textbf{Notations.} 
The quantized layers (including both linear and convolutional layers) in the mode are indexed by $l, 1\le l\le L$. The weights in the $l^{th}$ layer are denoted by $\mathbf{W}^{(l)}$ with  its vectorized/flattened version  $\mathbf{w}^{(l)}$.  $\mathbf{x}$ denotes its inputs. 

% \subsection{Quantization Framework}
% We first  quantize the polished activations with LogNP polishing  (Section~\ref{sec:activation_polishing}). Then we update weights to compensate the loss of activation quantization (Section~\ref{sec:compensation}). Finally, we quantize updated weights  (Section~\ref{sec:weight_quant}).

\subsection{Activation Polishing}\label{sec:activation_polishing}

Building on analysis in Section~\ref{sec:analysis_distribution}, we propose LogNP polishing method to adaptively smooth the deviant distribution in decoders in per-channel prospective.
For one activation $\mathbf x \in \mathbb{R}^{n \times d}$ where $n$ denotes the number of tokens and $d$ denotes the channel number, the LogNP polishing function $\Phi (\cdot, \cdot)$ for each element $x$ of input $\mathbf{x}$ is defined as follows,
\begin{equation}\label{eq:activation_polishing}
    \Phi({x}, \alpha) = \text{sign}({x}) \cdot \left[ \log_2 \left( |{x}| + \alpha \right) - \log_2(\alpha) \right],
\end{equation}
where $\text{sign}(\cdot)$ denotes the sign function,  and $\alpha$ is the polishing factor for one single channel.

\begin{figure}[t]
  \centering
  \includegraphics[width=1.0\linewidth]{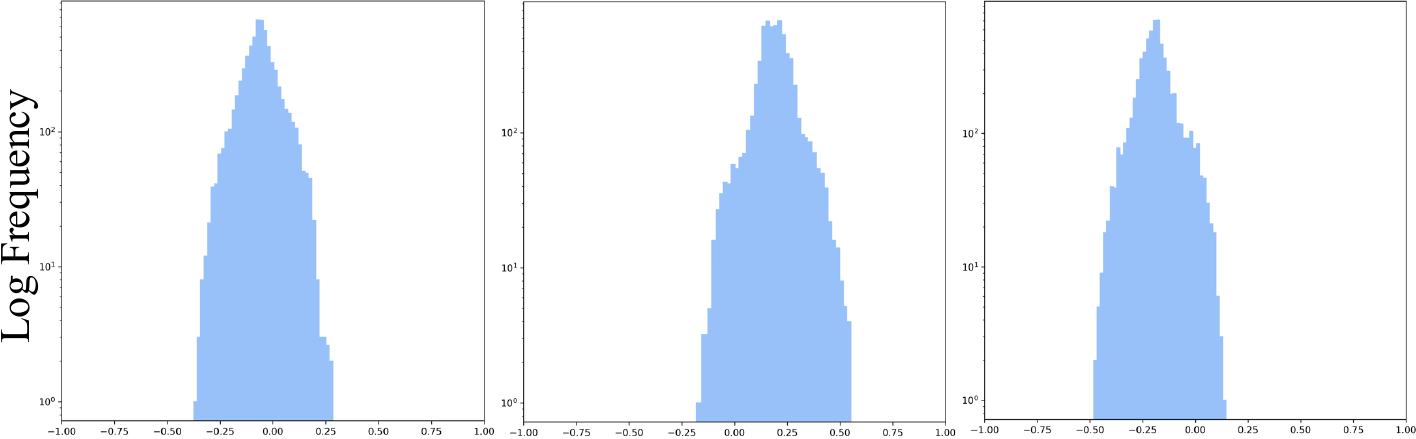}
  \caption{
  Visualization for the polished activation data distribution of different channels in decoder with $\log$ frequency.
  % \todo{todo: @changdi, insert a figure for the smooth of outlier deviant distribution, i.e., The polished activation colored in blue appears smoother compared to the original activation colored in red.}
  }
  \label{fig:method_activation_polishing}
\end{figure}

To reduce the quantization error in each channel, the polishing factor $\alpha_i$ corresponding to the $i^{th}$ channel with   activation $\mathbf{x}_i$  (i.e., $\mathbf{x} = \left[\mathbf{x}_1, \mathbf{x}_2, ..., \mathbf{x}_d\right] $) is computed according to the given percentile $\epsilon$ as follows,
%the polishing factor $\alpha = \left[\alpha_1, \alpha_2, ..., \alpha_d\right]$ for activation $\mathbf{x} = \left[\mathbf{x}_1, \mathbf{x}_2, ..., \mathbf{x}_d\right] $ is computed according to the given percentile $\epsilon$ as follows,
\begin{equation}\label{eq:activation_polishing_2}
    \alpha_i = P_{\epsilon}(\mathbf{x}_i), \ i \in \{1,2,3,...,d\}, \ \epsilon \in (0, 100)
\end{equation}
where $P_{\epsilon}(\cdot)$ denotes the $\epsilon$-th percentile function. We adopt uniform $\epsilon$ equals to 95 as deviant activation distributions remain similar across different inputs.
% We analyze deviant activation distributions in decoder and observe that they remain similar across different inputs. 
% Hence, we adopt uniform $\epsilon$ equals to 95 in our experiments.
% Therefore, in our experiments, $\epsilon$ is set to 95, and $\alpha$ is then calculated as the average 95th percentile value across the calibration samples.
We provide the visualization of the polished activation in Figure~\ref{fig:method_activation_polishing}.
The polished activation colored in blue appears smoother compared to the original activation colored in red shown in Figure~\ref{fig:analysis_outlier_distribution}, making it more friendly for quantization. After polishing, we perform activation quantization with uniform quantization in Section~\ref{sec:preliminary}.
Following activation dequantization, we adopt reverse transformation (i.e., unpolishing) as follows,
\begin{equation}
    \Phi^{-1}({\hat{x}}, \alpha) = \text{sign}({\hat{x}}) \cdot \left[ 2^{ \text{sign} ({\hat{x}}) \cdot \hat{x}  +  \log_2 \left( \alpha \right) } - \alpha \right]
\end{equation}
The latency overhead introduced by LogNP polishing and unpolishing is hidden by concurrent execution in our hardware design, which is further explained in Section~\ref{sec:on_chip_design}.

\subsection{Activation Loss Compensation} \label{sec:compensation}

The activation quantization introduces   quantization errors. To mitigate this issue, before weight quantization, we first  update all weights to compensate the loss of activation quantization. To save the computation cost, we address the problem layer by layer instead of for the whole model. The layer-wise compensation problem is formulated as the following,
\begin{align}
\min_{\Delta \mathbf W } \|  \mathbf W \mathbf{x} - (\mathbf W + \Delta \mathbf W  ) \hat{\mathbf{x}}  \|_2^2
\end{align}
where  $\mathbf W$ and  $\mathbf x$ are the original weights and inputs of the layer. $\hat{\mathbf{x}}$ is the quantized inputs and we update the weights with $\Delta \mathbf W $  to compensate the activation quantization error, so that  the layer outputs after activation quantization and weight compensation should be close to the original outputs. 

This problem can be solved by setting the gradients of the minimization objective to zero, as the following,
\begin{align}
\Delta \mathbf W   \hat{\mathbf{x}} \hat{\mathbf{x}}^{\text T} + \mathbf W( \mathbf{x} - \hat{\mathbf{x}}) \hat{\mathbf{x}}^{\text T} = 0 . 
\end{align}
The optimal solution can be obtained by
\begin{align}\label{eq:compensation_update}
\Delta \mathbf W^* = - \mathbf W( \mathbf{x} - \hat{\mathbf{x}}) \hat{\mathbf{x}}^{\text T} (\hat{\mathbf{x}} \hat{\mathbf{x}}^{\text T})^{-1}
\end{align}
After activation quantization, we can update weights with the above $\Delta \mathbf W^*$  to compensate the activation quantization error. If  $\hat{\mathbf{x}} \hat{\mathbf{x}}^{\text T}$ is not full rank with difficulties for matrix inversion, we adopt the dampening technique \cite{golub1965calculating,mottershead2006inverse}. Next, we quantize the updated weights under the quantized activations.  

\subsection{Weight Reconstruction}
\label{sec:weight_quant}

\textbf{Problem formulation.}
In quantization, the weights are modified and we would like to minimize the change of loss incurred by quantization  to  maintain the superior performance of the model.  To quantitatively analyze the loss degradation  by quantization, we approximate the loss degradation  with Taylor expansion  as  the following, 
\begin{align} \label{eq:H_loss}
    L(\mathbf{w}+\Delta\mathbf{w}) - L(\mathbf{w}) &\approx \Delta\mathbf{w}^{\text T}  \mathbf{g}_{\mathbf{w}} + \frac{1}{2}\Delta\mathbf{w}^{\text T} \mathbf{H}_{\mathbf{w}}\Delta\mathbf{w}, \nonumber \\
    &\approx  \frac{1}{2}\Delta\mathbf{w}^{\text T} \mathbf{H}_{\mathbf{w}}\Delta\mathbf{w},
\end{align}
where $\mathbf{g}_{\mathbf{w}}=\nabla_{\mathbf{w}}L$ represents the gradients and $\mathbf{H}_{\mathbf{w}}=\nabla^2_{\mathbf{w}}L$ denotes the Hessian matrix.  $\Delta\mathbf{w}$ is the weight perturbation in this step. Here we use the vectorized/flattened version $\mathbf w$ instead of the matrix version $\mathbf W$ to obtain a scalar output.  As the pre-trained model has been well trained and converged to a minimum,  it is reasonable to assume the gradients to  be $\mathbf{0}$.  The complexity to obtain the full Hessian is very high and typically it is not possible to obtain the full Hessian  due to computation and  memory limitations.  Specifically, $\mathbf{H}_{\mathbf{w}} \in \mathbb{R}^{B \times B}$  where  $B$ is the number of parameters in $\mathbf{w}$. The memory cost for $\mathbf{H}_{\mathbf{w}}$  is  extremely heavy in LLMs with billions of parameters. 

\noindent
\textbf{Approximating  Hessian.}
For a network trained with negative log-likelihood loss and a probabilistic model $p_{\mathbf{w}}(y | x_s)$ under softmax, the Hessian is identical to the Fisher matrix:
{\small\begin{align}
\label{eq:fim}
\mathbf{H}_{\mathbf{w}} = \mathbf{F}_{\mathbf{w}} = \sum\nolimits_{s=1}^S \mathbb{E}_{y\sim p_{\mathbf{w}}(y | x_s)} [& \nabla_{\mathbf{w}} \log p_{\mathbf{w}} (y | x_s) \nonumber \\ & \nabla_{\mathbf{w}} \log p_{\mathbf{w}} (y | x_s)^T ]
\end{align}}%
where $S$ is the number of samples, and for each sample, it needs to compute the expectation over all  categories.  The complexity with the expectation  is also high if there are too many categories. Thus, in practice, the  \textit{empirical Fisher} is commonly adopted to replace the expectation over $y$ with target label  $y_s$~\cite{kunstner2019limitations}.  The  dimension of $\mathbf{F}_{\mathbf{w}}$ is the same as $\mathbf{H}_{\mathbf{w}} $ with quadratic memory complexity to the number of parameters.  To reduce the complexity,  the empirical Fisher   is computed by    treating layers independently  and the layer-wise empirical Fisher can be obtained as below \cite{martens2015optimizing,botev2017practical}, 
\begin{align}
\mathbf{F}_{l}
&= \mathbb{E} \left[ (\mathbf{g}^{(l)} \mathbf{g}^{(l), \text T}) \otimes (\hat{\mathbf{x}}^{(l)} \hat{\mathbf{x}}^{(l), \text T}) \right],  
\end{align}
where $\hat{\mathbf{x}}^{(l)}$  is the input activations and $\mathbf{g}^{(l)} $ is the gradients of the $l^{th}$ layer. $\otimes$ denotes  Kronecker product.  The expectation is taken  with respect to  data distribution   over inputs.

For two different layers (for example, the $i^{th}$ and the $j^{th}$ layers), their Fisher is often written as 
$\mathbf{F}_{ij} = \sum_{s=1}^S \mathbb{E}\left[ \nabla_{\mathbf{w}^{(i)}} \log p_{\mathbf{w}}(y_s | x_s) \nabla_{\mathbf{w}^{(j)}} \log p_{\mathbf{w}}(y_s | x_s)^T \right]$, which are often   ignored to save computations. The Fisher for the whole model can be formulated as a block diagonal matrix 
$\mathbf{F}_{\mathbf{w}} =
\text{diag}(\mathbf{F}_{1}, \mathbf{F}_{2}, \ldots, \mathbf{F}_{L})$.

% positive semi-definite,

To obtain the Fisher in practice, we adopt  the KFAC approximation method \cite{martens2015optimizing,george2018fast} as below,  
% \begin{align}
% \mathbf{F}_{l}
% &= \sum\nolimits_{s=1}^S   (\mathbf{g}^{(l)}_{ s} \mathbf{g}_{ s}^{(l), \text T}) \otimes (\mathbf{a}^{(l)}_{s} \mathbf{a}^{(l), \text T}_{ s})  
% \end{align}
\begin{align}\label{eq:KFAC}
\mathbf{F}_l &= \mathbf{G}_l \otimes \mathbf{A}_l \\
\mathbf{G}_l &= \frac{1}{\sqrt{S}} \sum\nolimits_{s=1}^S 
\mathbf{g}_{ s}^{(l)} \mathbf{g}_{s}^{(l), \text T}    \\
\mathbf{A}_l &= \frac{1}{\sqrt{S}} \sum\nolimits_{s=1}^S \hat{\mathbf{x}}^{(l)}_{ s} \hat{\mathbf{x}}^{(l), \text T}_{ s}
\end{align}
The KFAC approximation \cite{wang2019eigendamage,immer2022invariance} approximates an expectation of Kronecker products as a Kronecker product of two expectations $\mathbb{E}[ \mathbf{g}^{(l)} \mathbf{g}^{(l),T} \otimes \hat{\mathbf{x}}^{(l)} \hat{\mathbf{x}}^{(l),T} ] \approx \mathbb{E}[ \mathbf{g}^{(l)} \mathbf{g}^{(l),T}] \otimes \mathbb{E}[\hat{\mathbf{x}}^{(l)} \hat{\mathbf{x}}^{(l),T} ] $, where the activations and derivatives are assumed to be independent.

\noindent
\textbf{Optimizing quantization parameters.}
Next we optimize the quantization parameters. Given our optimization loss in Equation (\ref{eq:H_loss}), multiple optimization methods can be adopted such as STE~\cite{hubara2018quantized} and AdaRound~\cite{adaround}. Due to its superior performance in PTQ, we use AdaRound~\cite{adaround} to learn the rounding parameter which decides rounding up or down. Formally, we minimize the following loss, 
{\small
\begin{align} \label{eq:F_loss}
\min_{ \mathbf{v}} \sum_{l=1}^{L}  \left( \mathbf{w}^{(l)}- \hat{\mathbf{w}}^{(l)}\right)^{\text T} \mathbf{F}_{l}  \left(\mathbf{w}^{(l)}- \hat{\mathbf{w}}^{(l)}\right) + \lambda h(\mathbf{v}),
\end{align}}%
where $\mathbf{v}$ denotes the rounding parameter which are used to construct the quantized weights  $\hat{\mathbf{w}}^{(l)}$. $h(\mathbf{v})$ is a  regularizer in AdaRound~\cite{adaround} with $\lambda$ controlling its strength.  More details about AdaRound are presented in Appendix \ref{app:sec:adaround}.
We only use a small calibration set (32 samples) sampled from  the  training dataset to calibrate the quantized model.

\begin{table*}[t]
\centering
\resizebox{1.0\linewidth}{!}{
\begin{tabular}{c|c|cccccc|cccccc}
\toprule
\multirow{2}{*}{Method} & \multirow{2}{*}{W / A} & \multicolumn{6}{c|}{NYUv2~\cite{nyuv2}}                                            & \multicolumn{6}{c}{KITTI~\cite{kitti}}                                            \\
\cline{3-14}
                        &                        & AbsRel $\downarrow$ & $\delta_1$ $\uparrow$ & $\delta_2$ $\uparrow$ & $\delta_3$ $\uparrow$ & RMSE $\downarrow$ & Silog $\downarrow$ & AbsRel $\downarrow$ & $\delta_1$ $\uparrow$ & $\delta_2$ $\uparrow$ & $\delta_3$ $\uparrow$ & RMSE $\downarrow$ & Silog $\downarrow$ \\
\midrule \multicolumn{14}{c}{ViT-Small Backbone}\\ \midrule
\textbackslash{}        & Float32               & 0.087          & 0.938          & 0.990          & 0.996          & 0.331          & 0.035          & 0.074          & 0.934          & 0.984          & 0.995          & 3.403          & 0.043          \\
\midrule
OBS~\cite{obs}                     & W4                     & 0.107          & 0.885          & 0.985          & 0.997          & 0.391          & 0.042          & 0.122          & 0.867          & 0.973          & 0.991          & 4.508          & 0.058          \\
\midrule
minmax~\cite{quant_minmax}                  & \multirow{6}{*}{W4A8}                   & 0.453          & 0.405          & 0.702          & 0.868          & 1.067          & 0.111          & 0.456          & 0.102          & 0.329          & 0.575          & 13.355         & 0.159          \\
ema~\cite{quant_ema}                     &                    & 0.257          & 0.591          & 0.859          & 0.958          & 0.793          & 0.093          & 0.312          & 0.341          & 0.632          & 0.807          & 11.102         & 0.139          \\
percentile~\cite{percentile}              &                    & 0.190          & 0.635          & 0.912          & 0.981          & 0.774          & 0.077          & 0.222          & 0.552          & 0.814          & 0.918          & 8.900          & 0.114          \\
AdaRound~\cite{adaround}&	&	0.134&	0.793&	0.967&	0.981&	0.471&	0.042&	0.128&	0.876&	0.953&	0.993&	4.236&	0.051\\
BrecQ~\cite{li2021brecq}                   &                    & 0.128          & 0.817          & 0.980          & 0.996          & 0.452          & 0.040          & 0.104          & 0.900          & 0.977          & 0.993          & 3.914          & 0.047          \\
\rowcolor{light-gray} Ours                    &                    & \textbf{0.103} & \textbf{0.899} & \textbf{0.987} & \textbf{0.996} & \textbf{0.380} & 0.042          & \textbf{0.092} & \textbf{0.916} & \textbf{0.981} & \textbf{0.995} & \textbf{3.682} & \textbf{0.046} \\
\midrule
minmax~\cite{quant_minmax}                  & \multirow{6}{*}{W4A4}                   & 1.265          & 0.130          & 0.285          & 0.477          & 2.593          & 0.160          & 0.584          & 0.054          & 0.160          & 0.321          & 15.877         & 0.245          \\
ema~\cite{quant_ema}                     &                    & 0.371          & 0.359          & 0.359          & 0.841          & 1.226          & 0.146          & 0.632          & 0.022          & 0.081          & 0.228          & 16.265         & 0.244          \\
percentile~\cite{percentile}              &                    & 0.357          & 0.366          & 0.659          & 0.846          & 1.223          & 0.146          & 0.646          & 0.019          & 0.064          & 0.197          & 16.384         & 0.244          \\
AdaRound~\cite{adaround}&	&	0.251&	0.459&	0.812&	0.936&	0.848&	0.103&	0.501&	0.239&	0.368&	0.538&	14.398&	0.216 \\
BrecQ~\cite{li2021brecq}                   &                    & 0.217          & 0.598          & 0.884          & 0.970          & 0.724          & 0.071          & 0.405          & 0.278          & 0.493          & 0.634          & 12.337         & 0.184          \\
\rowcolor{light-gray} Ours                    &                    & \textbf{0.133} & \textbf{0.815} & \textbf{0.965} & \textbf{0.990} & \textbf{0.453} & \textbf{0.052} & \textbf{0.197} & \textbf{0.534} & \textbf{0.937} & \textbf{0.985} & \textbf{5.429} & \textbf{0.059} \\
\midrule \multicolumn{14}{c}{ViT-Large Backbone}\\ \midrule
\textbackslash{}        & Float32               & 0.067          & 0.972          & 0.993          & 0.997          & 0.262          & 0.040          & 0.054          & 0.974          & 0.995          & 0.999          & 2.505          & 0.032          \\
\midrule
OBS~\cite{obs}                     & W4                     & 0.070          & 0.972          & 0.994          & 0.998          & 0.267          & 0.038          & 0.060          & 0.968          & 0.995          & 0.998          & 2.748          & 0.034          \\
\midrule
minmax~\cite{quant_minmax}                  & \multirow{6}{*}{W4A8}                   & 0.671          & 0.279          & 0.547          & 0.748          & 1.448          & 0.144          & 0.546          & 0.055          & 0.172          & 0.385          & 14.841         & 0.202          \\
ema~\cite{quant_ema}                     &                    & 0.657          & 0.289          & 0.559          & 0.756          & 1.425          & 0.141          & 0.394          & 0.219          & 0.481          & 0.686          & 12.085         & 0.169          \\
percentile~\cite{percentile}              &                    & 0.425          & 0.349          & 0.715          & 0.907          & 1.073          & 0.106          & 0.192          & 0.645          & 0.883          & 0.949          & 6.957          & 0.093          \\
AdaRound~\cite{adaround}&	&	0.084&	0.959&	0.991&	0.997&	0.276&	0.040&	0.057&	0.972&	0.995&	0.999&	2.769&	0.034 \\
BrecQ~\cite{li2021brecq}                   &                    & 0.076          & 0.967          & 0.993          & 0.997          & 0.272          & 0.039          & 0.057          & 0.971          & 0.995          & 0.999          & 2.778          & 0.034          \\
\rowcolor{light-gray} Ours                    &                    & \textbf{0.071} & \textbf{0.970} & \textbf{0.993} & \textbf{0.997} & \textbf{0.264} & \textbf{0.036} & \textbf{0.055} & \textbf{0.973} & \textbf{0.995} & \textbf{0.999} & \textbf{2.542} & \textbf{0.032} \\
\midrule
minmax~\cite{quant_minmax}                  & \multirow{6}{*}{W4A4}                   & 2.238          & 0.116          & 0.249          & 0.391          & 5.761          & 0.261          & 0.565          & 0.063          & 0.190          & 0.363          & 15.693         & 0.253          \\
ema~\cite{quant_ema}                     &                    & 1.039          & 0.174          & 0.376          & 0.578          & 2.126          & 0.157          & 0.508          & 0.123          & 0.294          & 0.460          & 15.206         & 0.247          \\
percentile~\cite{percentile}              &                    & 0.920          & 0.207          & 0.428          & 0.631          & 1.915          & 0.157          & 0.492          & 0.144          & 0.317          & 0.487          & 15.012         & 0.244          \\
AdaRound~\cite{adaround}&	&	0.501&	0.259&	0.564&	0.782&	1.482&	0.145&	0.482&	0.172&	0.395&	0.541&	13.782&	0.207 \\
BrecQ~\cite{li2021brecq}                   &                    & 0.469          & 0.330          & 0.609          & 0.811          & 1.323          & 0.137          & 0.434          & 0.191          & 0.407          & 0.590          & 13.462         & 0.192          \\
\rowcolor{light-gray} Ours                    &                    & \textbf{0.097} & \textbf{0.932} & \textbf{0.993} & \textbf{0.998} & \textbf{0.327} & \textbf{0.045} & \textbf{0.070} & \textbf{0.952} & \textbf{0.993} & \textbf{0.998} & \textbf{3.104} & \textbf{0.038} \\
\bottomrule
\end{tabular}}
  \caption{
  Main results of Metric3D model with ViT-Small and ViT-Large backbone on NYUv2 and KITTI datasets.
  Results with ViT-Giant backone is included in Table~\ref{tab:supp_results_metric3d_giant} at Appendix~\ref{sec:supp_additional_results}.
  }
  \label{tab:main_results_metric3d}
\end{table*}

\section{Hardware Design}\label{sec:asic_design}

% \todo{todo: @weize, please focus on the design of quantization, such as 4-bit kernel, logNP (i.e., Eq~\ref{eq:activation_polishing}) integration}

\subsection{Top-Level Design}

\begin{figure}[t]
  \centering
  \includegraphics[width=1.0\linewidth]{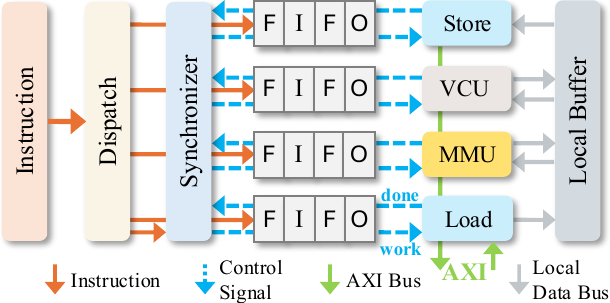}
  \caption{
  The block diagram of the proposed hardware accelerator.
  }
  \label{fig:top_hardware}
\end{figure}

We design a flexible and programmable accelerator as shown in Figure~\ref{fig:top_hardware}.
% The hardware consists of a Dispatch module for instruction fetch and dispatch, a Synchronize module for controlling the start and end of other modules, an Instruction FIFO for storing local instructions of each module, a Load and Store module for direct memory access(DMA), a matrix multiplication unit(MMU) for GEMM and convolution, a vector compute unit(VCU) for activation calculation and local buffers. 
%%%%%%%%%%%%%%% It comprises a Dispatch module for instruction fetching and dispatching, a Synchronize module for managing module execution, an Instruction FIFO for local instruction storage, a Load and Store module for DMA, a matrix multiplication unit (MMU) for GEMM and convolution, a vector compute unit (VCU) for activation calculations, and local buffers for data storage.
% The Dispatch module fetches instruction through the AXI bus and dispatches instruction into local Instruction FIFOs. The Synchronize module controls the start and the end of each module. Many out-of-order instructions are stored in the Instruction FIFOs of each module, controlled by the Synchronize module for launch and computation. The Load and Store module, which is the DMA of the hardware, transfers data between off-chip memory, like HBM or DDR, and local buffer. The MMU is responsible for performing convolution and matrix multiplication, while the VCU is used to compute nonlinear activation functions.
The Dispatch module fetches instructions via the AXI bus and distributes them to local Instruction FIFOs, while the Synchronize module controls the start and end of each module. 
Instruction FIFOs hold out-of-order instructions and are controlled by Synchronize module for execution. 
The Load and Store modules, functioning as the DMA, facilitate data transfer between off-chip memory (e.g., HBM or DDR) and local buffers.
The Matrix Multiplication Unit (MMU) performs General Matrix Multiply (GeMM) and convolution when the Vector Compute Unit (VCU) handles nonlinear activation functions.

\subsection{On-Chip Kernel Design}\label{sec:on_chip_design}
% In the Matrix Multiplication Unit (MMU), we implement a quantized matrix computation array to support integer GeMM.
% We adopt a multiply-accumulate tree array to perform convolution and matrix multiplication. 
% For different precision, we design corresponding multipliers and adders to enhance computational efficiency and reduce the overhead of chip area (e.g., for Float32, we implement IEEE 754 standard floating-point computation; for W4A4 or W4A8, we design customized integer multipliers and adders). 
% Benefit from the quantized matrix multiplication, the integer computation of W4A4 or W4A8 can greatly reduce the area overhead on hardware compared to Float32, so we can integrate more computational resources to obtain better performance.
% Furthermore, the low-bit quantized matrix alleviate the demand for memory bandwidth during data transfer, so the bandwidth of external memory can be fully utilized.
% % \todo{@weize, we can put such details of efficient GeMM design into supplementary}

We adopt a multiply-accumulate tree array to implement both matrix multiplication and convolution. 
For W4A4 and W4A8 configurations, we design specialized multipliers and adders to enhance computational efficiency and reduce chip area overhead. 
Additionally, the use of low-bit quantized matrices eases memory bandwidth demands during data transfer, allowing for full utilization of the external memory bandwidth.
As for Float32, we implement IEEE 754 standard floating-point computation for fair comparison.
% \todo{@weize, please write about: 1. the difference from other FPU; 2. compared to other quantization framework, what is your new design (difference)}
% For activation polishing optimization discussed in Section~\ref{sec:activation_polishing}, 

Meanwhile, we propose a programmable vector computation array composed of multiple Floating-Point Units (FPU) in Vector Computing Unit (VCU).
In the VCU, we design a programmable vector computation array composed of multiple FPU to enable more efficient computation. 
The FPU supports the Special Function Unit (SFU) through polynomial approximation~\cite{1388195}, which achieves the computation of non-linear functions (e.g. $\log_2$ and $\exp$) within 3 to 5 clock cycles while maintaining ultra-high precision. 
For the activation polishing optimization discussed in Section~\ref{sec:activation_polishing}, the FPU is further specifically designed to enhance execution efficiency.
Compared to the NVIDIA GPU~\cite{cuda_intrinsic}, the proposed SFU achieves nearly same accuracy, with only a 2 to 5 ULP (units in the last place) error.
Meanwhile, the VCU supports multiple numerical format inputs, unifying them into Float32 for computation, and then converts these Float32 numbers into integers such as INT4 or INT8 as needed, which seamlessly integrates both floating-point and integer data types for efficient processing.
Hence, the proposed quantization framework for MDE models with activation polishing optimization is accurately and efficiently deployed on our hardware.
% Compared to Intrinsic Functions in the SFU of NVIDIA GPU~\cite{cuda_intrinsic}, our SFU achieves nearly accuracy drop, with only 2 to 5 ulp error.
% Hence, the quantized GeMM with activation polishing is properly supported on our hardware.

% In the Vector Computing Unit (VCU), we design a programmable vector computation array composed of multiple Floating-Point Units (FPU) to support the quantization algorithms in Section~\ref{sec:activation_polishing}. 
% The FPU supports linear calculation (e.g. add and mul) in Float32, and a special function unit (SFU) using polynomial approximation can compute non-linear functions (e.g. log2 and exp2) within 3 to 5 clock cycles. 
% The VCU supports multiple numerical format inputs, unifying them into Float32 for computation, and then converts these Float32 numbers into integers such as INT4 or INT8 as needed. 
% As a result, the quantized GeMM and quantization algorithm in Section \ref{sec:activation_polishing} can be properly supported on our hardware.

\subsection{Overlapped Computation Flow}

% Our hardware architecture supports the kernel fusion of on-chip matrix multiplication and vector computation, where the intermediate result does not need to be written back to DDR and immediately sent to VCU for activation, quantization or dequantization. 
% Furthermore, our hardware is designed to be programmable using our customized instructions, allowing the configuration of different computational tasks to be executed concurrently within an instruction cycle. 
% As shown in Figure~\ref{fig:instruction}, data transferring, matrix computation, and vector computation can be fully overlapped to improve execution efficiency. 
% Therefore, the additional latency overhead introduced by the quantization algorithm in Section \ref{sec:activation_polishing} can be hidden due to the extensive operator fusion and concurrent execution, minimizing the impact on efficiency.

\begin{figure}[h]
  \centering
  \includegraphics[width=1.0\linewidth]{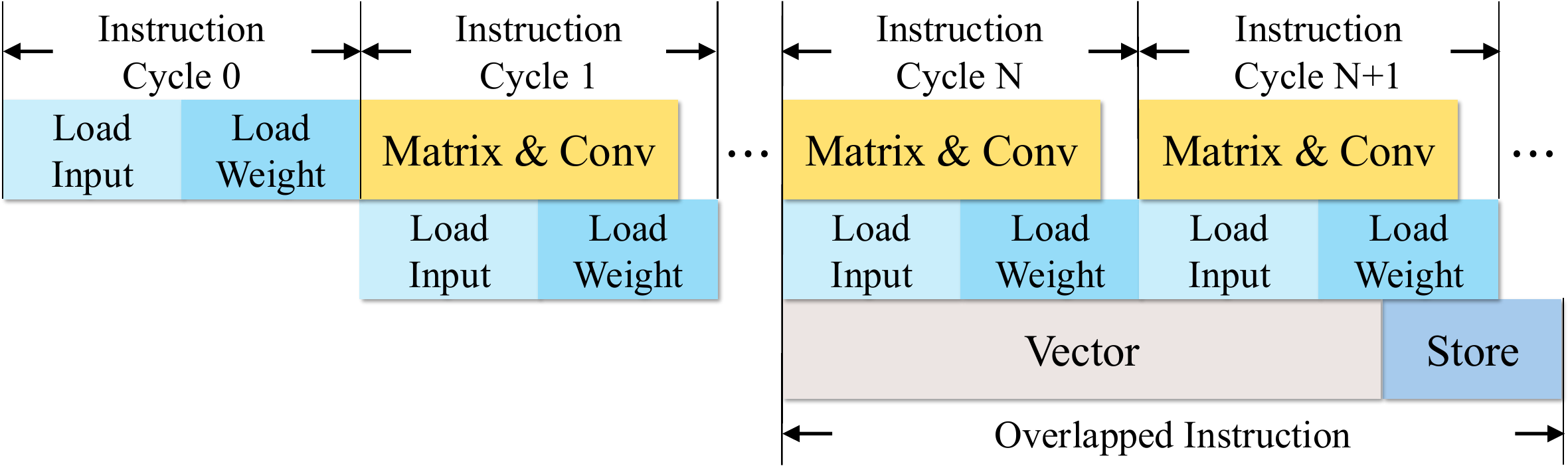}
  \caption{The schematic of concurrent execution, where the computation of Matrix, Vector, Load and Store can execute parallel. 
  % Matrix refers to matrix multiplication or convolution calculations that occur in the MMU, Vector refers to activation function calculations, quantization, and dequantization that occur in the VCU, and Load and Store refers to the data transfer between off-chip DDR and on-chip SRAM.
  % \todo{todo: @weize, this caption is duplication with section 5.4}
  }
  \label{fig:instruction}
\end{figure}

Our hardware architecture supports kernel fusion of on-chip matrix multiplication and vector computation, enabling intermediate results stored in local SRAM to be directly sent to the VCU for activation, quantization, or dequantization without being written back to DDR. 
By fusing these operations into a single flow, we minimize the latency introduced by multiple memory accesses.
% The latency introduced by multiple memory accesses is then reduced by the fusion of these operations.
Also, this design reduces data transfer time and enhances the overall execution efficiency.
Furthermore, our hardware is designed to be programmable with customized instructions, enabling the configuration of different computational tasks to be executed concurrently within an instruction cycle.
% This parallelism allows for a significant speedup in processing, as multiple operations can be pipelined, reducing the total time required to complete complex tasks.
This parallelism enables a substantial speedup in processing by allowing multiple operations to be pipelined, thereby reducing the total time needed to complete complex tasks.
As shown in Figure~\ref{fig:instruction}, data transferring, matrix computation, and vector computation can be fully overlapped to improve execution efficiency. 
This concurrency ensures that no computational resource is idle and  every cycle is utilized to its maximum potential, leading to a streamlined and optimized workflow.
% \todo{@weize, extend more details, before get the conclusion: 1. the difference of kernel fusion from others (how is the fusion when there are both floating and integer computation)}
Therefore, the total on-device latency of the depth estimation model executed with our quantization framework is optimized through extensive operator fusion and concurrent execution.

\begin{table*}[t]
\centering
\resizebox{1.0\linewidth}{!}{
\begin{tabular}{c|c|cc|cc|cc|cc|cc|cc|cc}
\toprule
\multirow{2}{*}{\rotatebox{90}{W / A}} & Dataset          & \multicolumn{2}{c|}{NYUv2~\cite{nyuv2}} & \multicolumn{2}{c|}{SUN RGB-D~\cite{sunrgbd}} & \multicolumn{2}{c|}{iBims-1~\cite{ibims1}} & \multicolumn{2}{c|}{HyperSim~\cite{hypersim}} & \multicolumn{2}{c|}{KITTI~\cite{kitti}} & \multicolumn{2}{c|}{vKITTI2~\cite{kitti}} & \multicolumn{2}{c}{DIODE Outdoor~\cite{diode}} \\
                         & Method           & AbsRel $\downarrow$     & $\delta_1$ $\uparrow$     & AbsRel $\downarrow$      & $\delta_1$ $\uparrow$      & AbsRel $\downarrow$     & $\delta_1$ $\uparrow$     & AbsRel $\downarrow$          & $\delta_1$ $\uparrow$         & AbsRel $\downarrow$     & $\delta_1$ $\uparrow$     & AbsRel $\downarrow$      & $\delta_1$ $\uparrow$      & AbsRel $\downarrow$          & $\delta_1$ $\uparrow$         \\
\midrule
\textbackslash{} & Float32 & 0.056        & 0.984      & 0.500         & 0.660       & 0.150        & 0.714      & 0.328             & 0.508          & 0.046        & 0.982      & 0.084         & 0.912       & 0.799             & 0.289          \\
\midrule
\rotatebox{90}{W4}                       & OBS~\cite{obs}              & 0.059        & 0.981      & 0.508         & 0.626       & 0.151        & 0.718      & 0.327             & 0.501          & 0.049        & 0.980      & 0.091         & 0.893       & 0.793             & 0.287          \\
\midrule
\multirow{2}{*}{\rotatebox{90}{W4A8}}    & BrecQ~\cite{li2021brecq}            & 0.099        & 0.903      & 0.489         & 0.692       & 0.183        & 0.593      & 0.377             & 0.372          & 0.051        & 0.951      & 0.606         & 0.132       & 0.799             & 0.010          \\
                         & \cellcolor{light-gray}Ours             & \cellcolor{light-gray}\textbf{0.058} & \cellcolor{light-gray}\textbf{0.982} & \cellcolor{light-gray}\textbf{0.394} & \cellcolor{light-gray}\textbf{0.756} & \cellcolor{light-gray}\textbf{0.157} & \cellcolor{light-gray}\textbf{0.700} & \cellcolor{light-gray}\textbf{0.322}   & \cellcolor{light-gray}\textbf{0.523}  & \cellcolor{light-gray}0.060 & \cellcolor{light-gray}\textbf{0.965} & \cellcolor{light-gray}\textbf{0.108} & \cellcolor{light-gray}\textbf{0.846} & \cellcolor{light-gray}\textbf{0.832}   & \cellcolor{light-gray}\textbf{0.298}          \\
\midrule
\multirow{2}{*}{\rotatebox{90}{W4A4}}    & BrecQ~\cite{li2021brecq}            & 0.342        & 0.296      & 0.396         & 0.416       & 0.419        & 0.222      & 0.703             & 0.047          & 0.385        & 0.235      & 0.757         & 0.009       & 0.765             & 0.034          \\
                         & \cellcolor{light-gray}Ours             & \cellcolor{light-gray}\textbf{0.070} & \cellcolor{light-gray}\textbf{0.972} & \cellcolor{light-gray}0.466 & \cellcolor{light-gray}\textbf{0.742} & \cellcolor{light-gray}\textbf{0.177} & \cellcolor{light-gray}\textbf{0.615} & \cellcolor{light-gray}\textbf{0.322}   & \cellcolor{light-gray}\textbf{0.512}  & \cellcolor{light-gray}\textbf{0.059} & \cellcolor{light-gray}\textbf{0.935} & \cellcolor{light-gray}\textbf{0.100} & \cellcolor{light-gray}\textbf{0.882} & \cellcolor{light-gray}\textbf{0.758}   & \cellcolor{light-gray}\textbf{0.280}         \\
\bottomrule
\end{tabular}}
  \caption{
  Main results of Depth Anything model with ViT-Large backbone. The first four sets are indoor scenes, while the last three are outdoor scenes. Full detailed results are included in Table~\ref{tab:supp_results_depth_anything_full} at Appendix~\ref{sec:supp_additional_results}.
  }
  \label{tab:main_results_depth_anything}
\end{table*}

\begin{table}[h]
\centering
\resizebox{1.0\linewidth}{!}{
\begin{tabular}{c|c|c|c|c|c}
\toprule
\multirow{2}{*}{W   / A} & Size                   & \multirow{2}{*}{Res.} & Latency & \multirow{2}{*}{FPS $\uparrow$} & Power Eff. \\
                         & (MB)                   &                       & (ms) $\downarrow$    &                      & (GMAC/W) $\uparrow$  \\
                         \midrule
\multicolumn{6}{c}{ViT-Small}                                                                                      \\
\midrule
Float32                  & 143.1                  & \multirow{3}{*}{256}  & 76.8 (1$\times$)    & 13.02 (1$\times$)               & 85.6 (1$\times$) \\
\cline{1-2} \cline{4-6}
W4A8                     & \multirow{2}{*}{17.88} &                       & 42.6 (1.8$\times$)    & 23.47 (1.8$\times$)               & 172.8 (2.0$\times$)  \\
W4A4                     &                        &                       & 38.3 (2.0$\times$)    & 26.11 (2.0$\times$)               & 177.5 (2.1$\times$) \\
\midrule
Float32                  & 143.1                  & \multirow{3}{*}{512}  & 325.5 (1$\times$)    & 3.07 (1$\times$)                 & 87.3 (1$\times$) \\
\cline{1-2} \cline{4-6}
W4A8                     & \multirow{2}{*}{17.88} &                       & 159.0 (2.0$\times$)   & 6.29 (2.0$\times$)                & 200.1 (2.3$\times$)  \\
W4A4                     &                        &                       & 133.4 (2.4$\times$)   & 7.50 (2.4$\times$)                 & 220.3 (2.5$\times$)  \\
\midrule
Float32                  & 143.1                  & \multirow{3}{*}{1024} & 2181.4 (1$\times$)  & 0.46 (1$\times$)                 & 78.4 (1$\times$) \\
\cline{1-2} \cline{4-6}
W4A8                     & \multirow{2}{*}{17.88} &                       & 836.8 (2.6$\times$)   & 1.20 (2.6$\times$)                 & 228.9 (2.9$\times$)  \\
W4A4                     &                        &                       & 616.7 (3.5$\times$)   & 1.62 (3.5$\times$)               & 287.0 (3.7$\times$)  \\
\midrule
\multicolumn{6}{c}{ViT-Large}                                                                                      \\
\midrule
Float32                  & 1572                   & \multirow{3}{*}{256}  & 584.1 (1$\times$)   & 1.71 (1$\times$)                 & 118.2 (1$\times$)  \\
\cline{1-2} \cline{4-6}
W4A8                     & \multirow{2}{*}{196.5} &                       & 223.5 (2.6$\times$)   & 4.47 (2.6$\times$)                 & 331.1 (2.8$\times$)  \\
W4A4                     &                        &                       & 167.9 (3.5$\times$)   & 5.96 (3.5$\times$)                 & 425.4 (3.6$\times$)  \\
\midrule
Float32                  & 1572                   & \multirow{3}{*}{512}  & 2214.5 (1$\times$)  & 0.45 (1$\times$)                 & 126.4 (1$\times$)  \\
\cline{1-2} \cline{4-6}
W4A8                     & \multirow{2}{*}{196.5} &                       & 798.4 (2.8$\times$)   & 1.25 (2.8$\times$)                 & 392.5 (3.1$\times$)  \\
W4A4                     &                        &                       & 568.5 (3.9$\times$)   & 1.76 (3.9$\times$)                 & 509.3 (4.0$\times$)  \\
\midrule
Float32                  & 1572                   & \multirow{3}{*}{1024} & 13297.1 (1$\times$) & 0.08 (1$\times$)                 & 126.9 (1$\times$)  \\
\cline{1-2} \cline{4-6}
W4A8                     & \multirow{2}{*}{196.5} &                       & 4228.3 (3.1$\times$)  & 0.24 (3.1$\times$)                 & 446.8 (3.5$\times$)  \\
W4A4                     &                        &                       & 2725.1 (4.9$\times$)  & 0.37 (4.9$\times$)                 & 640.6 (5.0$\times$)  \\
\bottomrule
\end{tabular}
}
  \caption{
  ASIC results of Metric3D with ViT-Small and ViT-Large. ViT-Giant results are shown in Table~\ref{tab:supp-latency_results_metric3d-8_core} at Appendix~\ref{sec:supp_additional_results}.
  }
  \label{tab:latency_results_metric3d-8_core}
\end{table}

% GMACs:
% small: 24.3242G, 105.1293G, 633.0681G
% large: 255.4823G 1.0356T 6.24386023T

\section{Experimental Results}

\subsection{Experiment Setup}

We adopt Metric3D~\cite{yin2023metric3d} with ViT-Small, ViT-Large, and ViT-Giant~\cite{vit} backbones to present the main results and Depth Anything~\cite{depth_anything_v1} with ViT-Large backbone to further verify the generalization across additional datasets.
For a fair comparison, we adopt per-channel asymmetric quantization for both weights and activations in W4A8 and W4A4 configurations.
We randomly sample 32 images as calibration set from the training sets of NYUv2~\cite{nyuv2} for indoor scenes and KITTI~\cite{kitti} for outdoor scenes.
The percentile $\epsilon$ for the polishing factor is set to 95, and $\alpha$ is then computed as the average value across the calibration samples.
For activation loss compensation and weight reconstruction, we still use same 32 samples for the calibration.
For gradient computation in weight reconstruction, we adopt batch size of 1, learning rate set to 4e-5, warm up of 0.2, weight decay of 0.01, drop rate of 0.5, and 20,000 iterations.
For evaluation, we use NYUv2~\cite{nyuv2}, SUN RGB-D~\cite{sunrgbd}, iBims-1~\cite{ibims1}, and HyperSim~\cite{hypersim} for indoor scenes; and KITTI~\cite{kitti}, vKITTI~\cite{vkitti2}, and DIODE~\cite{diode} for outdoor scenes.
Absolute relative error (AbsRel), accuracy under threshold ($\delta_i < 1.25^i, i=1,2,3$), root mean squared error (RMSE), and metric-depth loss (Silog)  metrics are employed in our evaluation results.

\subsection{Hardware Implementation} 
% \todo{@weize, please write about testing environment}
We implement our hardware in RTL and synthesis the design using Design Compiler~\cite{DesignCompiler} under a commercial 28nm CMOS technology. 
The frequency and area are tested by the Design Compiler and the power is tested by Prime Time~\cite{PrimeTime} PX with synthesized netlist and dynamic simulation switch rate under different modes.
After synthesis, based on the latency profiled by simulation and the frequency given by Design Compiler, we introduce a hardware performance emulator for latency test. The DDR is simulated by RTL, with a bandwidth of 19.2 GB/s.
% Latency result are tested by the performance emulator.

\subsection{Depth Estimation Results}

We present the quantization results with W4A8 and W4A4 configurations using Metric3D~\cite{yin2023metric3d} with ViT-Small and ViT-Large backbone on indoor dataset NYUv2~\cite{nyuv2} and outdoor dataset KITTI~\cite{kitti} in Table~\ref{tab:main_results_metric3d}. 
The results with ViT-Giant backbone are included in Table~\ref{tab:supp_results_metric3d_giant} at Appendix~\ref{sec:supp_additional_results}. 
Compared to conventional PTQ methods such as minmax~\cite{quant_minmax}, EMA~\cite{quant_ema}, and percentile~\cite{percentile}, our framework demonstrates significantly improved performance. Additionally, when benchmarked against other learning-based approaches including OBS~\cite{obs}, AdaRound~\cite{adaround}, and BrecQ~\cite{li2021brecq}, our method also achieves superior results. Notably, for the ViT-Large backbone with W4A8 settings, our method maintains nearly lossless performance. In the W4A4 configuration, our approach shows a substantial performance advantage over other methods, validating the effectiveness of our proposed quantization framework.
Meanwhile, we present results of Depth Anything model with ViT-Large backbone on various additional datasets~\cite{sunrgbd, ibims1, hypersim, vkitti2, diode} covering both indoor and outdoor scenes, as shown in Table~\ref{tab:main_results_depth_anything}.
Our method, which applies quantization to both weights and activations, achieves performance comparable to OBS (a weight-only quantization method) across multiple datasets in W4A8 configuration, and outperforms the BrecQ for all configurations, showing the effectiveness of our proposed quantization method. 
% Notably, in W4A8 configuration, our method even outperforms the FP32 baseline. One possible reason is that \todo{Need to find a reason}

% \todo{@pu, please give more acc results analysis}

\subsection{Latency Results on ASIC}

We present the latency results of Metric3D with detailed hardware resource utility in Table~\ref{tab:latency_results_metric3d-8_core}.
The latency is tested with three different resolutions.
The power is further normalized by the latency and computation cost (i.e., GMACs) for the energy efficiency.
% \todo{@weize, please write how power and area is tested (according to GMACs)}
For three different configurations, the frequency are uniformly 1 GHz.
The silicon area that our design occupies on a chip for Float32, W4A8, and W4A4 is 29.22 mm$^2$, 23.94 mm$^2$, and 24.35 mm$^2$, respectively.
As shown in Table~\ref{tab:latency_results_metric3d-8_core}, our proposed method demonstrates superior performance with faster inference speed and higher power efficiency across various model scales and resolutions. 
Notably, for ViT-Small backbone at 256 resolution under W4A4 configuration, our method achieves up to 26 FPS, showing the real-time performance.
% Notably, for ViT-Large at 1024 resolution in the W4A4 configuration, our method achieves impressive gains, delivering a 4.9× speedup in inference and a 5× improvement in power efficiency.
% \todo{@pu, please extend}

% \input{tables/main_results-depth_anything-vit_large}

% hardware main table
% model size, latency (FPS), power, area, frequency
% resolution 256x256
% Float32
% W4A8
% W4A4
% resolution 512x512
% Float32
% W4A8
% W4A4
% resolution 1024x1024
% Float32
% W4A8
% W4A4

% \input{tables/latency_results-metric3d-8_core}

\subsection{Ablation Study}

\begin{figure}[t]
  \centering
  \includegraphics[width=1.0\linewidth]{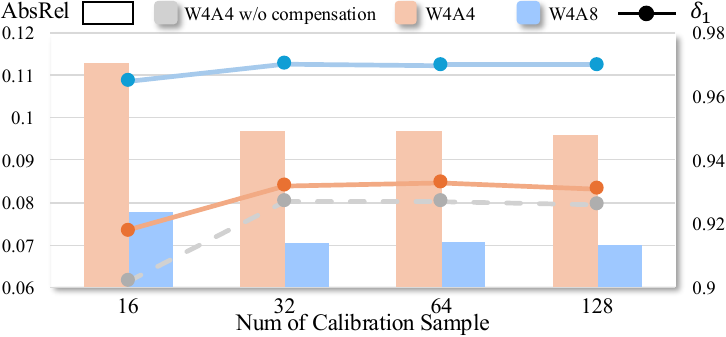}
  \caption{
  Accuracy ablation for the number of calibration samples using ViT-Large backbone on NYUv2 dataset.
  }
  \label{fig:ablation_calibration_sample}
\end{figure}

We regulate the number of calibration samples to verify the effectiveness of our proposed method, and the results are shown in Figure~\ref{fig:ablation_calibration_sample}.
We observe that the model performance stabilizes when the number of calibration samples reaches 32 for both W4A4 and W4A8 configuration. 
Additionally, activation compensation proves crucial for the W4A4.

% ablation calibration sample figure 
% todo: ablation for with or without activation loss compensation

\begin{figure}[t]
  \centering
  \includegraphics[width=1.0\linewidth]{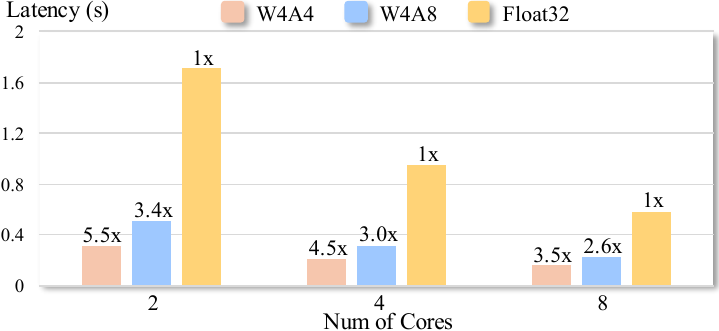}
  \caption{
  Latency ablation for different number of cores using ViT-Large backbone with 256x256 resolution.
  }
  \label{fig:ablation_latency_cores}
\end{figure}

% We further ablate the hardware resource with different number of MMUs and VCUs to resize the computation ability
% number of kernels \todo{@weize, please say how to achieve this ablation.}
We further perform ablation study on hardware resources by varying the number of MMUs and VCUs (i.e., number of cores).
% to adjust and evaluate the computational capability.
% 2 cores represent a configuration containing 2 MMUs and 2 VCUs. 
The results of 2, 4 and 8 cores are presented in Figure~\ref{fig:ablation_latency_cores}.
The results show that more cores lead to faster speed.  Compared to the Float32 model, the quantized model is not sensitive to the hardware resource  and achieves high hardware efficiency.

% ablation hardware resource figure
% 2, 4, 8-core, ViT-Large backbone
% latency (FPS), power, area, frequency

% \subsection{Qualitative Results}

% To exhibit the superiority of our proposed Quart-Depth framework, especially on low-bit quantization (W4A4), 
Meanwhile, we also visualize depth estimation results of Metric3D with ViT-Large backbone in Figure~\ref{fig:visualization} at Appendix~\ref{sec:supp_visual_results} compared with BrecQ~\cite{li2021brecq} on NYUv2~\cite{nyuv2} for indoor scene and KITTI~\cite{kitti} for outdoor scene.
The visualizations verify the effectiveness of our methods on both indoor and outdoor scenes.
% \todo{@pu, extend if needed}

% \begin{figure}[t]
%   \centering
%   \includegraphics[width=1.0\linewidth]{figures/visualization.pdf}
%   \caption{
%   Visualization of the quantized Metric3D (W4A4) with ViT-Large backbone for indoor scenes (top two rows) and outdoor scenes (bottom two rows).
% %  in  configuration. The top two rows denote indoor scenes and the bottom two rows denote the outdoor scenes.
%   }
%   \label{fig:visualization}
% \end{figure}

% \todo{@changdi, get some visualizations}

\section{Conclusion}

In this paper, we propose  \textbf{QuartDepth}, a quantization framework for the real-time acceleration of MDE models on  ASICs. 
We provide the LogNP polishing method to smooth the outliers in the decoder of MDE models, and provide the compensation method to mitigate the loss of activation quantization. 
We then introduce the weight reconstruction to reduce the weight quantization loss. 
Furthermore, we design a novel flexible and programmable hardware accelerator for our efficient deployment  on ASICs.
Experimental results verify the effectiveness of our proposed framework. 
% \todo{@pu, please extend if necessary}

\section{Acknowledgment}
This work was mainly supported by Northeastern University and Nanjing University. This work was also supported by National Natural Science Foundation of China 62472104.

{
    \small
    \bibliographystyle{ieeenat_fullname}
    \bibliography{reference}
}

\clearpage
\setcounter{page}{1}
\maketitlesupplementary

\setcounter{table}{0}
\renewcommand{\thetable}{A\arabic{table}}
\renewcommand*{\theHtable}{\thetable}

\setcounter{figure}{0}
\renewcommand{\thefigure}{A\arabic{figure}}
\renewcommand*{\theHfigure}{\thefigure}

\section{Quantization with AdaRound} \label{app:sec:adaround}

We adopt adaptive rounding (AdaRound)  \cite{adaround} for weight quantization as it performs well in post-training quantization. Specifically, different from traditional quantization with rounding-to-nearest operation, AdaRound optimizes the rounding policy  so that all weights can learn the final rounding. All weights are initially rounded by floor operation. Then  a learnable variable $\mathbf{v}$ is trained to determine the final rounding policy (i.e., flooring or ceiling) for each weight. The formulation  can be  given by
\begin{equation}
    \hat{\mathbf{w}} = s\times\mathrm{clip}\left(\lfloor\frac{\mathbf{w}}{s} \rfloor
    + \sigma(\mathbf{v})+zp, 0, 2^k-1\right). 
    \label{eq_adaround}
\end{equation}
$\mathbf{v}$ is the learnable parameter and 
the sigmoid-like function $\sigma(\cdot)$ keeps the learnable variable $\mathbf{v}$ moving between 0 and 1.  The  loss in quantization is formulated as following,
{\small
\begin{align} \label{app:eq:F_loss}
\min_{ \mathbf{v}} \sum_{l=1}^{L}  \left( \mathbf{w}^{(l)}- \hat{\mathbf{w}}^{(l)}\right)^{\text T} \mathbf{F}_{l}  \left(\mathbf{w}^{(l)}- \hat{\mathbf{w}}^{(l)}\right) + \lambda h(\mathbf{v}),
\end{align}}%
where
\begin{equation}
     h(\mathbf{v}) = \sum_i \left(1-|2\sigma(\mathbf{v}_i)-1|^{\beta}\right).
\end{equation}
We have a  regularization term $h(\mathbf{v})$ in the  loss to ensure that  $\sigma(\mathbf{v})$ converges to either 0 or 1 with a decreasing $\beta$.

\section{Additional Results}\label{sec:supp_additional_results}

\paragraph{Metric3D ViT-Giant Backbone Results.}
We further present the results of Metric3D~\cite{yin2023metric3d} model with ViT-Giant backbone in Table~\ref{tab:supp_results_metric3d_giant} on NYUv2~\cite{nyuv2} and KITTI~\cite{kitti} datasets. 
The results show that our method achieves superior performance compared to other quantization methods especially with W4A4 configuration.

\begin{table*}[h]
\centering
\resizebox{1.0\linewidth}{!}{
\begin{tabular}{c|c|cccccc|cccccc}
\toprule
\multirow{2}{*}{Method} & \multirow{2}{*}{W / A} & \multicolumn{6}{c|}{NYUv2~\cite{nyuv2}}                                            & \multicolumn{6}{c}{KITTI~\cite{kitti}}                                            \\
\cline{3-14}
                        &                        & AbsRel $\downarrow$ & $\delta_1$ $\uparrow$ & $\delta_2$ $\uparrow$ & $\delta_3$ $\uparrow$ & RMSE $\downarrow$ & Silog $\downarrow$ & AbsRel $\downarrow$ & $\delta_1$ $\uparrow$ & $\delta_2$ $\uparrow$ & $\delta_3$ $\uparrow$ & RMSE $\downarrow$ & Silog $\downarrow$ \\
\midrule \multicolumn{14}{c}{ViT-Giant Backbone}\\ \midrule
\textbackslash{}        & Float32               & 0.071       & 0.970     & 0.994     & 0.998     & 0.266   & 0.029    & 0.061       & 0.974     & 0.995     & 0.999     & 2.431   & 0.030    \\
\midrule
OBS~\cite{obs}                     & W4                     & 0.088       & 0.936     & 0.994     & 0.999     & 0.275   & 0.038    & 0.072       & 0.967     & 0.995     & 0.999     & 2.672   & 0.036    \\
\midrule
minmax~\cite{quant_minmax}                  & \multirow{6}{*}{W4A8}                   & 0.942       & 0.068     & 0.263     & 0.588     & 2.063   & 0.100    & 0.212       & 0.559     & 0.832     & 0.953     & 8.058   & 0.110    \\
ema~\cite{quant_ema}                     &                    & 0.867       & 0.098     & 0.326     & 0.645     & 1.850   & 0.101    & 0.219       & 0.533     & 0.820     & 0.954     & 8.069   & 0.110    \\
percentile~\cite{percentile}              &                    & 0.926       & 0.079     & 0.282     & 0.597     & 1.986   & 0.098    & 0.231       & 0.501     & 0.786     & 0.940     & 8.325   & 0.112    \\
AdaRound~\cite{adaround}&	&	0.140&	0.871&	0.995&	0.999&	0.358&	0.034&	0.093&	0.961&	0.994&	0.998&	2.694&	0.033 \\
BrecQ~\cite{li2021brecq}                   &                    & 0.141       & 0.867     & 0.995     & 0.999     & 0.362   & 0.034    & 0.093       & 0.956     & 0.994     & 0.998     & 2.710   & 0.033    \\
\rowcolor{light-gray} Ours                    &                    & \textbf{0.093} & \textbf{0.941} & \textbf{0.995} & \textbf{0.999} & \textbf{0.272} & 0.035 & \textbf{0.076} & \textbf{0.965} & \textbf{0.995} & \textbf{0.999} & \textbf{2.687} & \textbf{0.031}    \\
\midrule
minmax~\cite{quant_minmax}                  & \multirow{6}{*}{W4A4}                   & 1.844       & 0.062     & 0.145     & 0.287     & 3.808   & 0.161    & 0.407       & 0.277     & 0.481     & 0.632     & 13.911  & 0.237    \\
ema~\cite{quant_ema}                     &                    & 1.943       & 0.050     & 0.126     & 0.253     & 3.999   & 0.156    & 0.389       & 0.312     & 0.534     & 0.683     & 13.370  & 0.236    \\
percentile~\cite{percentile}              &                    & 2.039       & 0.043     & 0.112     & 0.228     & 4.219   & 0.154    & 0.385       & 0.317     & 0.553     & 0.710     & 12.909  & 0.231    \\
AdaRound~\cite{adaround}&	&	0.737&	0.032&	0.359&	0.810&	1.976&	0.061&	0.186&	0.591&	0.968&	0.994&	4.591&	0.045 \\
BrecQ~\cite{li2021brecq}                   &                    & 0.749       & 0.032     & 0.339     & 0.791     & 1.977   & 0.063    & 0.186       & 0.583     & 0.968     & 0.994     & 4.711   & 0.045    \\
\rowcolor{light-gray} Ours                    &                    & \textbf{0.119} & \textbf{0.901} & \textbf{0.994} & \textbf{0.999} & \textbf{0.333} & \textbf{0.037} & \textbf{0.068} & \textbf{0.958} & \textbf{0.993} & \textbf{0.998} & \textbf{2.938} & \textbf{0.036}    \\
\bottomrule
\end{tabular}}
  \caption{
  Results of Metric3D~\cite{yin2023metric3d} model with ViT-Giant backbone on NYUv2 and KITTI datasets.
  }
  \label{tab:supp_results_metric3d_giant}
\end{table*}

\paragraph{Depth Anything Full Results.}
We present the detailed results of Depth Anything~\cite{depth_anything_v1} model in Table~\ref{tab:supp_results_depth_anything_full} with additional evaluation metrics on multiple datasets including indoor and outdoor scenes. 
The results show that our method achieves better performance than other two methods on nearly all evaluation metrics, especially with W4A4 configuration.

\begin{table*}[h]
\centering
\resizebox{0.77\linewidth}{!}{
\begin{tabular}{c|c|ccccccccc}
\toprule
\multicolumn{2}{c|}{Dataset}              & \multicolumn{9}{c}{NYUv2~\cite{nyuv2}}                                                             \\
\midrule
Method           & W / A                 & AbsRel $\downarrow$ & $\delta_1$ $\uparrow$ & $\delta_2$ $\uparrow$ & $\delta_3$ $\uparrow$ & RMSE $\downarrow$   & log10 $\downarrow$ & RMSElog $\downarrow$ & Silog $\downarrow$   & SqRel $\downarrow$ \\
\midrule
\textbackslash{} & Float32      & 0.056    & 0.984  & 0.998  & 1.000  & 0.206  & 0.024   & 0.072     & 5.277  & 0.017   \\
\midrule
OBS~\cite{obs}              & W4                    & 0.059    & 0.981  & 0.998  & 1.000  & 0.214  & 0.025   & 0.075     & 5.478  & 0.018   \\ \midrule
BrecQ~\cite{li2021brecq}            & W4A8 & 0.099    & 0.903  & 0.994  & 0.999  & 0.439  & 0.046   & 0.128     & 8.563  & 0.059   \\
\rowcolor{light-gray} Ours             & W4A8                      & \textbf{0.058} & \textbf{0.982} & \textbf{0.998} & \textbf{1.000} & \textbf{0.214} & \textbf{0.025} & \textbf{0.075} & \textbf{5.461}  & \textbf{0.018} \\
\midrule
BrecQ~\cite{li2021brecq}            & W4A4 & 0.342    & 0.296  & 0.596  & 0.826  & 1.264  & 0.179   & 0.472     & 30.389 & 0.500   \\
\rowcolor{light-gray} Ours             & W4A4                      & \textbf{0.070} & \textbf{0.972} & \textbf{0.997} & \textbf{0.999} & \textbf{0.268} & \textbf{0.031} & \textbf{0.090} & \textbf{7.071}  & \textbf{0.025} \\
\midrule
\multicolumn{2}{c|}{Dataset}              & \multicolumn{9}{c}{SUN RGB-D~\cite{sunrgbd}}                                                         \\
\midrule
% Method           & W / A                 & abs\_rel & delta1 & delta2 & delta3 & rmse   & log\_10 & rmse\_log & silog  & sq\_rel \\
\textbackslash{} & Float32      & 0.500    & 0.660  & 0.960  & 0.980  & 0.616  & 0.088   & 0.259     & 15.483 & 2.175   \\
\midrule
OBS~\cite{obs}              & W4                    & 0.508    & 0.626  & 0.959  & 0.980  & 0.624  & 0.091   & 0.266     & 15.337 & 2.141   \\ \midrule
BrecQ~\cite{li2021brecq}            & W4A8 & 0.489    & 0.692  & 0.961  & 0.981  & 0.597  & 0.084   & 0.252     & 15.498 & 2.149   \\
\rowcolor{light-gray} Ours             & W4A8                      & \textbf{0.394} & \textbf{0.756} & \textbf{0.962} & \textbf{0.983} & \textbf{0.447} & \textbf{0.076} & \textbf{0.232} & \textbf{15.086} & \textbf{1.211} \\
\midrule
BrecQ~\cite{li2021brecq}            & W4A4 & 0.396    & 0.416  & 0.702  & 0.868  & 0.816  & 0.151   & 0.416     & 31.762 & 0.426   \\
\rowcolor{light-gray} Ours             & W4A4                      & \textbf{0.466} & \textbf{0.742} & \textbf{0.962} & \textbf{0.981} & \textbf{0.554} & \textbf{0.079} & \textbf{0.241} & \textbf{15.759} & 2.053          \\
\midrule
\multicolumn{2}{c|}{Dataset}              & \multicolumn{9}{c}{iBims-1~\cite{ibims1}}                                                           \\
\midrule
% Method           & W / A                 & abs\_rel & delta1 & delta2 & delta3 & rmse   & log\_10 & rmse\_log & silog  & sq\_rel \\
\textbackslash{} & Float32      & 0.150    & 0.714  & 0.966  & 0.991  & 0.593  & 0.073   & 0.185     & 7.515  & 0.130   \\
\midrule
OBS~\cite{obs}              & W4                    & 0.151    & 0.718  & 0.968  & 0.991  & 0.598  & 0.073   & 0.185     & 7.549  & 0.130   \\ \midrule
BrecQ~\cite{li2021brecq}            & W4A8 & 0.183    & 0.593  & 0.941  & 0.980  & 0.764  & 0.092   & 0.227     & 7.679  & 0.195   \\
\rowcolor{light-gray} Ours             & W4A8                      & \textbf{0.157} & \textbf{0.700} & \textbf{0.958} & \textbf{0.986} & \textbf{0.628} & \textbf{0.077} & \textbf{0.194} & 7.741           & \textbf{0.144} \\
\midrule
BrecQ~\cite{li2021brecq}            & W4A4 & 0.419    & 0.222  & 0.410  & 0.621  & 1.899  & 0.261   & 0.665     & 29.776 & 0.979   \\
\rowcolor{light-gray} Ours             & W4A4                      & \textbf{0.177} & \textbf{0.615} & \textbf{0.952} & \textbf{0.989} & \textbf{0.696} & \textbf{0.088} & \textbf{0.219} & \textbf{8.049}  & \textbf{0.168} \\
\midrule
\multicolumn{2}{c|}{Dataset}              & \multicolumn{9}{c}{HyperSim~\cite{hypersim}}                                                          \\
\midrule
% Method           & W / A                 & abs\_rel & delta1 & delta2 & delta3 & rmse   & log\_10 & rmse\_log & silog  & sq\_rel \\
\textbackslash{} & Float32      & 0.328    & 0.508  & 0.709  & 0.824  & 3.370  & 0.166   & 0.421     & 15.999 & 1.893   \\
\midrule
OBS~\cite{obs}              & W4                    & 0.327    & 0.501  & 0.706  & 0.821  & 3.407  & 0.169   & 0.427     & 15.961 & 1.897   \\ \midrule
BrecQ~\cite{li2021brecq}            & W4A8 & 0.377    & 0.372  & 0.629  & 0.764  & 3.894  & 0.205   & 0.509     & 17.426 & 2.199   \\
\rowcolor{light-gray} Ours             & W4A8                      & \textbf{0.322} & \textbf{0.523} & \textbf{0.717} & \textbf{0.830} & \textbf{3.347} & \textbf{0.163} & \textbf{0.414} & \textbf{15.776} & \textbf{1.869} \\
\midrule
BrecQ~\cite{li2021brecq}            & W4A4 & 0.703    & 0.047  & 0.103  & 0.179  & 6.291  & 0.549   & 1.317     & 34.756 & 4.684   \\
\rowcolor{light-gray} Ours             & W4A4                      & \textbf{0.322} & \textbf{0.512} & \textbf{0.713} & \textbf{0.828} & \textbf{3.381} & \textbf{0.166} & \textbf{0.422} & \textbf{15.999} & \textbf{1.859} \\
\midrule
\multicolumn{2}{c|}{Dataset}              & \multicolumn{9}{c}{KITTI~\cite{kitti}}                                                             \\
\midrule
% Method           & W / A                 & abs\_rel & delta1 & delta2 & delta3 & rmse   & log\_10 & rmse\_log & silog  & sq\_rel \\
\textbackslash{} & Float32      & 0.046    & 0.982  & 0.998  & 1.000  & 1.897  & 0.020   & 0.069     & 6.106  & 0.121   \\
\midrule
OBS~\cite{obs}              & W4                    & 0.049    & 0.980  & 0.998  & 0.999  & 1.971  & 0.021   & 0.072     & 6.325  & 0.137   \\ \midrule
BrecQ~\cite{li2021brecq}            & W4A8 & 0.051    & 0.951  & 0.985  & 0.997  & 3.349  & 0.041   & 0.082     & 7.237  & 0.328   \\
\rowcolor{light-gray} Ours             & W4A8                      & 0.060          & \textbf{0.965} & \textbf{0.996} & \textbf{0.999} & \textbf{2.687} & \textbf{0.027} & 0.090          & 8.217           & \textbf{0.205} \\
\midrule
BrecQ~\cite{li2021brecq}            & W4A4 & 0.385    & 0.235  & 0.429  & 0.632  & 12.683 & 0.359   & 1.258     & 8.391  & 3.682   \\
\rowcolor{light-gray} Ours             & W4A4                      & \textbf{0.059} & \textbf{0.935} & \textbf{0.981} & \textbf{0.989} & \textbf{4.194} & \textbf{0.056} & \textbf{0.093} & \textbf{7.622}  & \textbf{0.425} \\
\midrule
\multicolumn{2}{c|}{Dataset}              & \multicolumn{9}{c}{vKITTI2~\cite{vkitti2}}                                                           \\
\midrule
% Method           & W / A                 & abs\_rel & delta1 & delta2 & delta3 & rmse   & log\_10 & rmse\_log & silog  & sq\_rel \\
\textbackslash{} & Float32      & 0.084    & 0.912  & 0.986  & 0.995  & 4.008  & 0.039   & 0.138     & 12.096 & 0.430   \\
\midrule
OBS~\cite{obs}              & W4                    & 0.091    & 0.893  & 0.981  & 0.994  & 4.489  & 0.042   & 0.149     & 12.894 & 0.517   \\ \midrule
BrecQ~\cite{li2021brecq}            & W4A8 & 0.606    & 0.132  & 0.275  & 0.447  & 18.052 & 0.734   & 1.721     & 31.342 & 11.434  \\
\rowcolor{light-gray} Ours             & W4A8                      & \textbf{0.108} & \textbf{0.846} & \textbf{0.968} & \textbf{0.992} & \textbf{5.436} & \textbf{0.052} & \textbf{0.175} & \textbf{13.870} & \textbf{0.720} \\
\midrule
BrecQ~\cite{li2021brecq}            & W4A4 & 0.757    & 0.009  & 0.024  & 0.065  & 18.789 & 0.731   & 1.856     & 77.917 & 11.753  \\
\rowcolor{light-gray} Ours             & W4A4                      & \textbf{0.100} & \textbf{0.882} & \textbf{0.979} & \textbf{0.994} & \textbf{4.461} & \textbf{0.047} & \textbf{0.156} & \textbf{13.128} & \textbf{0.546} \\
\midrule
\multicolumn{2}{c|}{Dataset}              & \multicolumn{9}{c}{DIODE Outdoor~\cite{diode}}                                                     \\
\midrule
% Method           & W / A                 & abs\_rel & delta1 & delta2 & delta3 & rmse   & log\_10 & rmse\_log & silog  & sq\_rel \\
\textbackslash{} & Float32      & 0.799    & 0.289  & 0.611  & 0.837  & 6.641  & 0.187   & 0.531     & 34.917 & 9.447   \\
\midrule
OBS~\cite{obs}              & W4                    & 0.793    & 0.287  & 0.604  & 0.830  & 6.685  & 0.188   & 0.534     & 34.929 & 9.130   \\ \midrule
BrecQ~\cite{li2021brecq}            & W4A8 & 0.799    & 0.010  & 0.024  & 0.053  & 13.264 & 0.690   & 1.629     & 36.942 & 9.932   \\
\rowcolor{light-gray} Ours             & W4A8                      & 0.832          & \textbf{0.298} & \textbf{0.618} & \textbf{0.844} & \textbf{6.632} & \textbf{0.186} & \textbf{0.532} & \textbf{35.208} & 10.520         \\
\midrule
BrecQ~\cite{li2021brecq}            & W4A4 & 0.765    & 0.034  & 0.081  & 0.146  & 13.575 & 0.695   & 1.706     & 58.540 & 10.101  \\
\rowcolor{light-gray} Ours             & W4A4                      & \textbf{0.758} & \textbf{0.280} & \textbf{0.605} & \textbf{0.825} & \textbf{6.801} & \textbf{0.189} & \textbf{0.533} & \textbf{35.280} & \textbf{8.470} \\
\bottomrule
\end{tabular}}
  \caption{
  Full results of Depth Anything~\cite{depth_anything_v1} model with ViT-Large backbone.
  }
  \label{tab:supp_results_depth_anything_full}
\end{table*}

\paragraph{Latency Results with ViT-Giant Backbone.}
We provide the latency results with ViT-Giant backbone in Table~\ref{tab:supp-latency_results_metric3d-8_core}. 
The results show that our quantized model achieves faster inference and higher power efficiency compared to the Float32 model. 
Particularly with W4A4 configuration, our method achieves 5.3$\times$ faster inference and 5.5$\times$ power efficiency.

\begin{table}[h]
\centering
\resizebox{1.0\linewidth}{!}{
\begin{tabular}{c|c|c|c|c|c}
\toprule
\multirow{2}{*}{W   / A} & Size                   & \multirow{2}{*}{Res.} & Latency  & \multirow{2}{*}{FPS $\uparrow$} & Power Eff. \\
                         & (MB)                   &                       & (ms) $\downarrow$    &                      & (GMAC/W)$\uparrow$   \\
                         \midrule
\multicolumn{6}{c}{ViT-Giant}                                                                                      \\
\midrule
Float32                  & 5258.5                 & \multirow{3}{*}{256}  & 1769.6 (1$\times$)  & 0.57 (1$\times$)                & 122.1 (1$\times$) \\
\cline{1-2} \cline{4-6}
W4A8                     & \multirow{2}{*}{656.9} &                       & 610.4 (2.9$\times$)  & 1.64 (2.9$\times$)                & 396.6 (3.2$\times$) \\
W4A4                     &                        &                      & 424.4 (4.2$\times$)  & 2.36 (4.2$\times$)                & 527.0 (4.3$\times$) \\
\midrule
Float32                  & 5258.5                 & \multirow{3}{*}{512}  & 6446.7 (1$\times$)  & 0.16 (1$\times$)                 & 134.1 (1$\times$)  \\
\cline{1-2} \cline{4-6}
W4A8                     & \multirow{2}{*}{656.9} &                       & 2133.4 (3.0$\times$)  & 0.47 (3.0$\times$)                 &  419.4 (3.1$\times$) \\
W4A4                     &                        &                       &  1425.8 (4.5$\times$) &  0.70 (4.5$\times$)                & 679.1 (5.1$\times$) \\
\midrule
Float32                  & 5258.5                 & \multirow{3}{*}{1024} & 36308.1 (1$\times$)  & 0.03 (1$\times$)                 & 142.9 (1$\times$)  \\
\cline{1-2} \cline{4-6}
W4A8                     & \multirow{2}{*}{656.9} &                       & 10992.7 (3.3$\times$) & 0.09 (3.3$\times$)                & 528.5 (3.7$\times$) \\
W4A4                     &                        &                       & 6790.3 (5.3$\times$)  & 0.15 (5.3$\times$)                & 790.5 (5.5$\times$) \\
\bottomrule
\end{tabular}}
  \caption{
  Latency results of Metric3D with ViT-Giant backbone.
  }
  \label{tab:supp-latency_results_metric3d-8_core}
\end{table}

\section{Visualization Results}\label{sec:supp_visual_results}

\begin{figure}[h]
  \centering
  \includegraphics[width=1.0\linewidth]{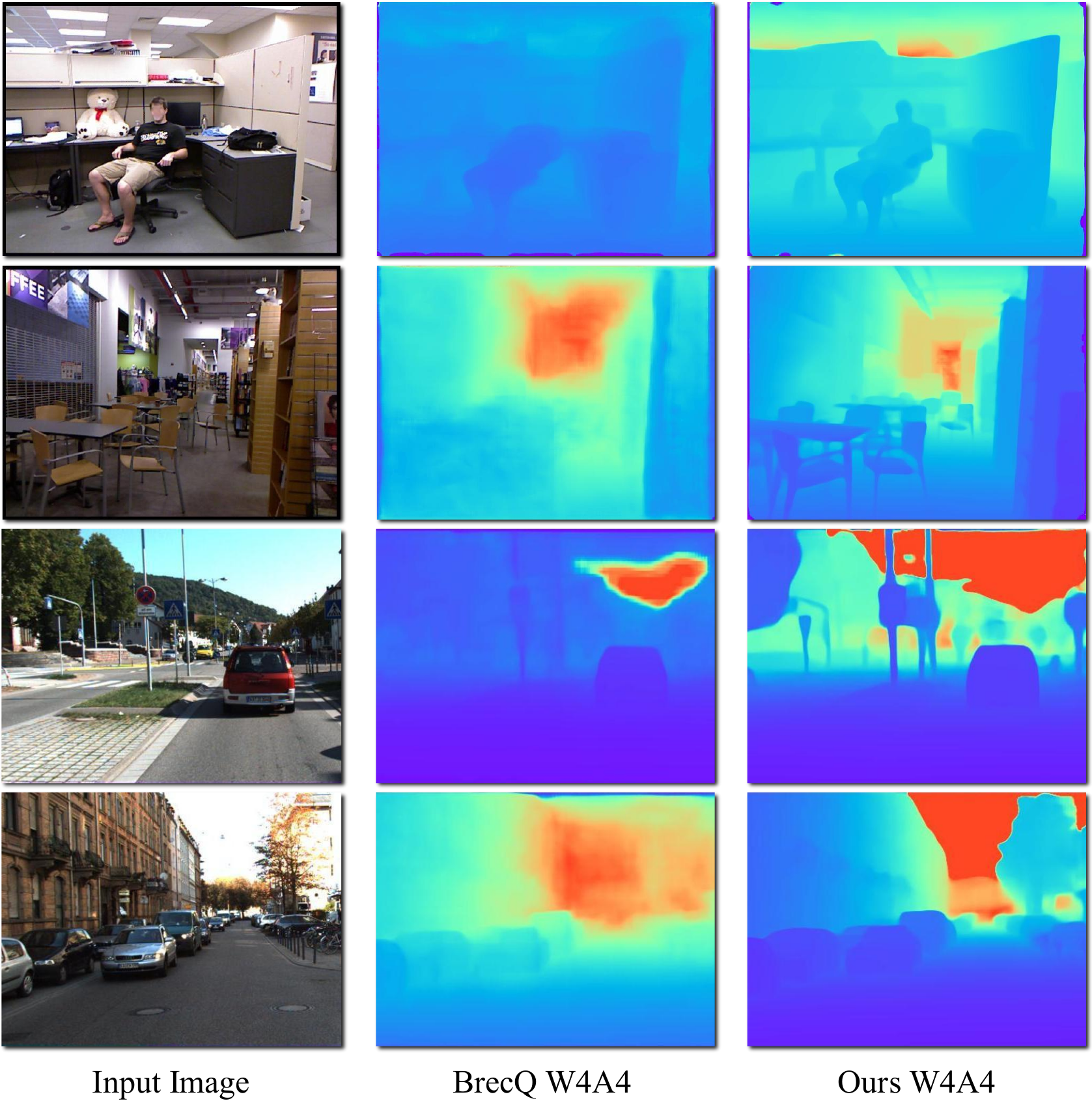}
  \caption{
  Visualization of the quantized Metric3D (W4A4) with ViT-Large backbone for indoor scenes (top two rows) and outdoor scenes (bottom two rows).
%  in  configuration. The top two rows denote indoor scenes and the bottom two rows denote the outdoor scenes.
  }
  \label{fig:visualization}
\end{figure}

\end{document}